\documentclass{article}

\usepackage{microtype}
\usepackage{graphicx}
\usepackage{subfigure}
\usepackage{booktabs} 
\usepackage{nicefrac}       
\usepackage{microtype}      

\usepackage{hyperref}
\usepackage{lipsum}
\usepackage{comment}
\usepackage{multirow}

\usepackage[accepted]{icml2024}

\usepackage{amsfonts}       
\usepackage{amsmath}
\usepackage{amssymb}
\usepackage{mathtools}
\usepackage{amsthm}
\usepackage{caption}
\usepackage{enumitem}
\usepackage{caption}
\usepackage{wrapfig}
\usepackage{graphicx}
\usepackage{subfigure}
\usepackage{physics}

\usepackage{pifont}
\newcommand{\cmark}{\ding{51}}%
\newcommand{\xmark}{\ding{55}}%

\usepackage{hyperref}
\usepackage[capitalize,noabbrev]{cleveref}

\theoremstyle{plain}

\theoremstyle{definition}

\theoremstyle{remark}

\usepackage[textsize=tiny]{todonotes}

\icmltitlerunning{Sparse Iso-FLOP Transformations for Maximizing Training Efficiency}

\begin{document}

\twocolumn[
\icmltitle{Sparse-IFT: Sparse Iso-FLOP Transformations for \\ Maximizing Training Efficiency}



\icmlsetsymbol{equal}{*}
\icmlsetsymbol{work}{$\dagger$}

\begin{icmlauthorlist}
\icmlauthor{Vithursan Thangarasa}{equal,yyy}
\icmlauthor{Shreyas Saxena}{equal,work}
\icmlauthor{Abhay Gupta}{work}
\icmlauthor{Sean Lie}{yyy}
\end{icmlauthorlist}

\icmlaffiliation{yyy}{Cerebras Systems Inc, California, USA, ${}^{\dagger}$Work done while at Cerebras}

\icmlcorrespondingauthor{Vithursan Thangarasa}{vithu@cerebras.net}

\icmlkeywords{Sparsity, Efficient Training, Machine Learning, ICML}

\vskip 0.3in
]



\printAffiliationsAndNotice{\icmlEqualContribution} 

\begin{abstract}
Recent research has focused on weight sparsity in deep neural network training
to reduce FLOPs, aiming for improved efficiency (test accuracy~w.r.t training
FLOPs). However, sparse weight training often compromises accuracy, requiring
extended training schedules to attain the accuracy of dense models. In contrast,
our approach, Sparse Iso-FLOP Transformations (Sparse-IFT), uses sparsity to
\textit{improve accuracy} while maintaining dense model FLOPs. Using a single
hyperparameter (i.e., the sparsity level), Sparse-IFTs efficiently replace dense
layers, expanding the search space for optimal sparse masks. In addition,
dynamic sparse training (DST) with Sparse-IFT models effectively navigate this
larger sparse mask-weight space, which is evidenced by a spectral analysis using
Ramanujan graph properties. Our study reveals a robust correlation among mask
topology, weights, and final performance. Notably, without adjusting any
training hyperparameters, replacing dense layers with Sparse-IFT yields
significant improvements, such as a +3.5\% boost for ResNet-18 on ImageNet and
+0.9\% for GPT-3 Small on the Open LLM leaderboard. To the best of our
knowledge, this is the first work to demonstrate the use of sparsity for
improving the accuracy of dense models through a set of simple-to-use sparse
transformations. Code is available
at:~\url{https://github.com/CerebrasResearch/Sparse-IFT}.
\end{abstract}
\section{Introduction}

\begin{figure}[ht]
    \vspace{-5pt}
    \begin{center}
    \centerline{\includegraphics[width=1.0\columnwidth]{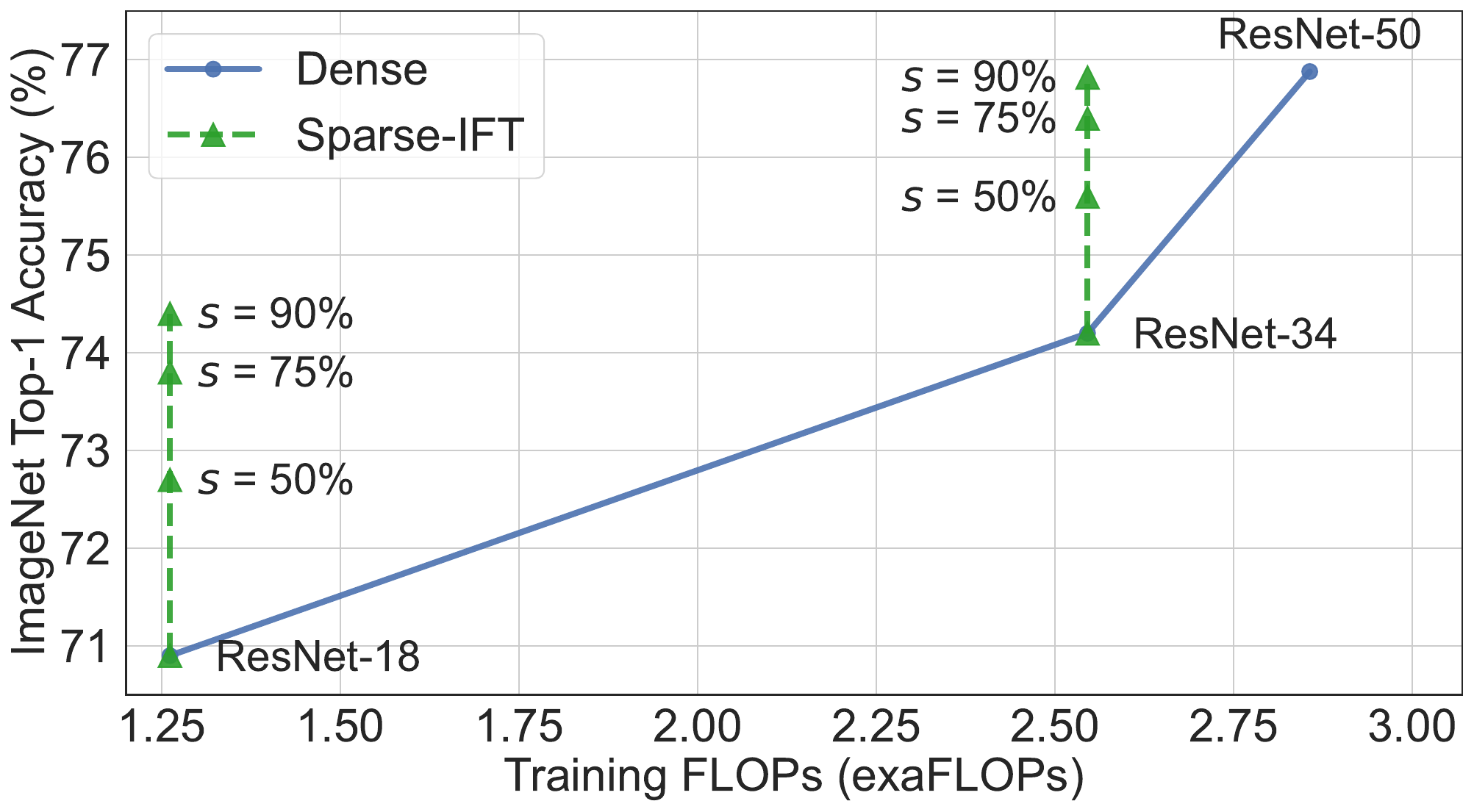}}
    \vspace{-5pt}
    \caption{Top-1 Accuracy vs. Training FLOPs for variants of ResNet on
    ImageNet. Sparse-IFT provides significant accuracy gains across different
    models and sparsity levels, $s \in \{\text{50\%, 75\%, 90\%}\}$, while using the same training FLOPs as its dense
    counterpart. }
    \label{fig:sift_resnet_improvement}
    \end{center}
    \vspace{-30pt}
\end{figure}

Increases in model size and training data have led to many breakthroughs in deep
learning (e.g., AlexNet~\citep{krizhevsky2012imagenet},
ResNet~\citep{he2016identity}, Transformers~\citep{vaswani2017attention},
GPT~\citep{radford2018improving, radford2019language},
AlphaGo~\citep{silver2017mastering}, etc.). Consequently, computational and
memory demands for training and deploying deep neural networks (DNNs) have
surged dramatically. To enable the deployment of large models, multiple
techniques (e.g., distillation~\citep{hinton2015distilling},
quantization~\citep{han2015deep}, pruning~\citep{han2015learning}) have been
introduced to reduce inference FLOPs and memory requirements. While these
techniques improve inference efficiency (test accuracy w.r.t inference FLOPs),
the associated training costs are still prohibitive. Our work focuses on
improving the training efficiency (test-accuracy w.r.t training FLOPs) of DNNs
through weight sparsity. In recent years, we have witnessed progress in using
weight sparsity to reduce training FLOPs~\citep{evci2020rigging, liu2021sparse,
jayakumar2020top}. \citet{frankle2018lottery} show that sparse subnetworks
(``lottery tickets'') exist at initialization and can be trained to match dense
network accuracy. Dynamic sparse training (DST) methods~\citep{ma2022effective,
evci2020rigging, liu2021selfish, jayakumar2020top} iteratively adjust sparsity
patterns to facilitate the discovery of optimal sparse subnetworks within a
single training run. However, they often lag behind dense baselines or require
longer training schedules (e.g., 2-5x training steps) to close the
gap~\citep{yuan2021mest, tai2022spartan, liu2021sparse}. Our unique contribution
focuses on using sparsity to improve a given dense model's accuracy. We
introduce the Sparse Iso-FLOP Transformations (Sparse-IFT), a family of
techniques serving as drop-in replacements for dense layers in DNNs.

Sparse-IFTs increase layer representational capacity, facilitating the discovery
of optimal sparse subnetworks while maintaining constant FLOPs (i.e., Iso-FLOP).
For example, widening a layer with maintained sparsity increases dimensionality
without impacting FLOPs; expanding the sparse mask-weight space for more diverse
configurations. This enables DST methods to navigate the search space
effectively, potentially finding improved sparse subnetworks for higher
accuracy. Drawing inspiration from prior works~\citep{hoang2023revisiting,
hoang2023dont}, we analyze the connectivity of Sparse-IFT models as Ramanujan
graphs and their impact on performance when trained with DST. All Sparse-IFTs
are parameterized by a single hyperparameter, the sparsity level.
Figure~\ref{fig:sift_resnet_improvement} summarizes ImageNet performance,
showing significant accuracy gains with Sparse Wide IFT ResNet variants. Sparse
Wide ResNet-18 achieves +3.5\% top-1 accuracy at 90\% sparsity, surpassing a
dense ResNet-34 (74.2\%) with 2x fewer FLOPs. These gains result from replacing
dense layers with Sparse-IFTs, requiring no changes to training hyperparameters.
The main contributions of our work are:
\vspace{-5pt}
\begin{enumerate}
    \item We introduce Sparse Iso-FLOP Transformations (Sparse-IFTs), a family
    of techniques aimed at enhancing DNN training efficiency. These
    transformations boost accuracy while maintaining a constant FLOP count.
    Sparse-IFTs are parameterized by a \textit{single hyperparameter, sparsity
    level}, and can be seamlessly used as drop-in replacements for dense layers.
    \item We empirically validate the consistent advantage of DST over static
    sparse training for Sparse-IFT networks. Our investigation into the dynamic
    evolution of sparse topologies in DST via Ramanujan graph spectral analysis
    highlights optimized connectivity patterns and improved spectral
    characteristics.

    \item We show consistent benefits of Sparse-IFT across computer vision and
    natural language processing domains. Sparse-IFT enhances ResNet-18 and
    ResNet-34 top-1 accuracy on ImageNet by 3.5\% and 2.6\%, respectively.
    Fine-tuning for object detection (MS COCO) and segmentation (CityScapes)
    yields improvements of 5.2\% mAP and 2.4\% mIoU. Sparse-IFT with GPT-3
    results in a 0.9\% improvement on the Open LLM leaderboard.

    \item We showcase the practical value of Sparse-IFT with real-world timings
    for training on the Cerebras CS-2~\citep{lie2023cerebras}
    and inference with Neural Magic DeepSparse~\citep{neural_magic_2021} using
    unstructured sparsity. Despite being 2x wider at 75\% sparsity with Sparse
    Wide IFT, we observe minimal compute overhead on both platforms compared to
    GPUs.

\end{enumerate}
\vspace{-10pt}
\section{Method}
\label{sec:method}

\begin{figure*}[h]
\includegraphics[width=\textwidth]{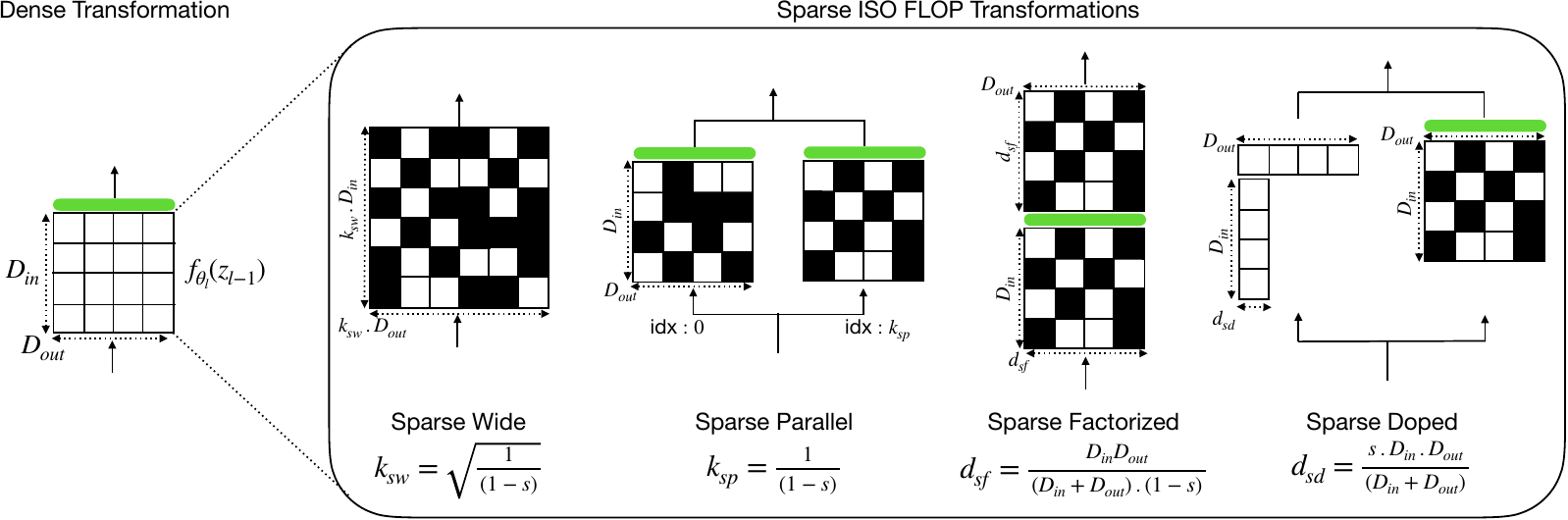}

\vspace{-0.05in}
\caption{Different members of the Sparse-IFT family, each parameterized by a
single hyperparameter (i.e., sparsity level, $s$). Black and white squares
denote non-active and active weights, respectively. Green block indicates a
non-linear activation function (e.g., ReLU). Derived with sparsity set at $50\%$
as an example, all transformations are Iso-FLOP to the dense feedforward
function $f_{\theta_l}$, making them suitable drop-in replacements for
$f_{\theta_l}$. Details about each member are in
Section~\ref{subsec:members_of_sift}.}
\label{fig:diff_members_of_sift}
\vspace{-10pt}
\end{figure*}

In this section, we first explain our intuition and hypotheses, followed by our
methodology to improve training efficiency.


\vspace{-10pt}
\paragraph{Training with Dense Matrices is FLOP Inefficient}
Modern DNNs are often overparameterized, showing sparsity in features and
weights across layers. The Lottery Ticket Hypothesis~\cite{frankle2018lottery,
chen2020lottery} suggests sparse DNNs, initialized with an effective sparsity
mask (``lottery ticket''), can achieve the same accuracy as dense counterparts.
Sparse training methods theoretically enhance efficiency but often yield lower
accuracy than dense baselines. This discrepancy may stem from challenges in
identifying optimal masks within a single training run. Existing sparse training
methods~\cite{jayakumar2020top, evci2020rigging, yuan2021mest, tai2022spartan,
liu2021sparse} invest these FLOP savings into longer training schedules to
bridge accuracy gaps, inefficiently requiring more FLOPs than dense baselines
for the same target accuracy.

In our work, we take an orthogonal approach and invest these FLOP savings to (1)
enhance a layer's representational capacity and (2) expand its search space,
aiming to discover an optimal sparse mask~\cite{ramanujan2020s,
stosic2021search}. Larger sparse models show potential for improved accuracy,
but designing an appropriate architecture is challenging. For instance,
achieving performance surpassing ResNet-18 on ImageNet requires careful balance
of sparsity and network size. Existing studies explore diverse combinations but
often lack FLOP efficiency, requiring multiple iterations for optimal settings
and hyperparameter tuning. To address this, we propose the Sparse Iso-FLOP
Transformation (Sparse-IFT) family, replacing dense layers with FLOP-equivalent
sparse transformations. Notably, Sparse-IFT is parameterized by a single
hyperparameter—the sparsity level, simplifying the tuning process.

\vspace{-5pt}
\subsection{Sparse Iso-FLOP Transformations}
\paragraph{Setup} For clarity, we explain our method in the context of a fully
connected network.
Let $\mathcal{N}$ denote a $L$ layered DNN parameterized by
$\Theta_{\mathcal{N}}$. Let $\Theta_{\mathcal N} \in \{\theta_1, \ldots,
\theta_L\}$ denote the parameters of the DNN. The output of the $l$-th layer is
defined as: $z_l = \sigma (f_{\theta_l}(z_{l-1}))$ for some activation function
$\sigma$ (e.g., ReLU~\cite{nair2010rectified}) and feedforward function
$f_{\theta_l}$. Specifically, let $f_{\theta_l}(z_{l-1}) = \theta_l^T z_{l-1}$,
where $\theta_l \in \mathbb{R}^{D_{in} \times D_{out}}$,  $z_{l-1} \in
\mathbb{R}^{D_{in} \times B}$ and $B$, $D_{in}$, $D_{out}$ denote the
batch-size, input, and output dimensionality of features respectively. The total
FLOPs needed for $f_{\theta_l}$ are given by $B{\cdot}D_{in}{\cdot}D_{out}$.  In
Appendix~\ref{app:sift_conv}, we detail a straightforward extension to
convolutional layers. In the standard setup, the feedforward function
$f_{\theta_l}$ computes output features through a linear transformation of input
features. While theoretically, arbitrary non-linear transformations can be
applied, practical implementations often resort to expressing transformations as
dense matrix multiplications for efficient GPU
support~\citep{nvidia2023gpuperf}. We aim to boost DNN training efficiency by
enhancing the representational capacity of the feedforward function. Unlike
conventional methods that increase capacity by stacking more
layers~\cite{lin2014network}, widening~\cite{zagoruyko2016wide}, or
ensembling~\citep{etai2020colleg}, our approach introduces unstructured sparsity
in weight matrices, achieving the same FLOPs as a dense feedforward function. 

Let $\Psi_l$ denote the set of Sparse Iso-FLOP Transformations (Sparse-IFT) for
a particular layer $l$: \setlength{\belowdisplayskip}{2pt}
\setlength{\belowdisplayshortskip}{2pt} \setlength{\abovedisplayskip}{2pt}
\setlength{\abovedisplayshortskip}{2pt}
\begin{equation*}
\Psi_l: \{ \psi_l(s),  0 \leq s < 1,  g(\psi_l) \approx g(f_{\theta_l}) \},
\end{equation*}
where $\psi_l$ is a transformation, $s$ represents the sparsity level, and
$g(\cdot)$ returns the computational FLOPs. Each transformation in this set
satisfies the following properties: (1) the computational FLOPs of the
transformation $\psi_l$ are same as that of dense transformation $f_{\theta_l}$,
and (2) the transformation is parameterized by a single hyperparameter - the
sparsity level. These Iso-FLOP transformations serve as drop-in replacements for
dense feedforward functions, preserving layer FLOPs. While there may be other
FLOP-invariant transformations, in this work, we explore: Sparse Wide, Sparse
Parallel, Sparse Factorized, and Sparse Doped.

\vspace{-5pt}
\subsection{Members of Sparse-IFT}
\label{subsec:members_of_sift}
\paragraph{Sparse Wide}
This transformation augments the representational capacity of a layer by
increasing the number of output features while keeping $s$ fraction of weights
sparse. Hence, it widens the input and output features for all $L$ layers of the
network with the same widening factor, $k_{sw}$, to avoid mismatch in feature
dimensionality across layers. Let $\theta_l^{sw} \in
\mathbb{R}^{k_{sw}{\cdot}D_{in} \times k_{sw}{\cdot}D_{out}}$ denote the
transformation matrix, with $s$ fraction of weights being sparse. Since the
fraction of non-sparse weights is given by $1-s$, the FLOPs required by this
transformation are
$B{\cdot}(k_{sw}{\cdot}D_{in}){\cdot}(k_{sw}{\cdot}D_{out}){\cdot}(1-s)$.
Setting these equal to the FLOPs of the original dense $f_{\theta_l}$, we obtain
the widening factor $k_{sw} = \sqrt{1/(1-s)}$. If we set the sparsity $s$ to
$0$, we obtain $k_{sw}$ as $1$ and recover the  dense feedforward function.
\vspace{-15pt}
\paragraph{Sparse Parallel} The sparse parallel transformation replaces the
feedforward function with a sum of $k_{sp}$ non-linear functions. Let
$\theta_l^{sp} \in \{ \theta_l^{sp, 1},\ldots, \theta_l^{sp, k_{sp}}\}$ denote
the parameters of this transformation, where $\theta_l^{sp, j} \in
\mathbb{R}^{D_{in}\times D_{out}}$ denotes the transformation matrix of $j^{th}$
function, where $s$ fraction of weights are sparse. The sparse parallel
transformation in this case is $\psi^{sp}_l =
\sum^{k_{sp}}_{j=1}\sigma((\theta_l^{sp, j})^{T} z_{l})$, where $\sigma$ is a
non linear function. In practice, $\psi^{sp}_l$ is implemented as a layer with
$k_{sp}$ parallel branches. The computational FLOPs of this transformation is
$k_{sp}{\cdot}B{\cdot}D_{in}{\cdot}D_{out}{\cdot}(1-s)$. Setting these FLOPs
equal to FLOPs of $f_{\theta}$, we obtain $k_{sp} = 1/(1-s)$. Note, at $s=0$,
the number of parallel branches $k_{sp}$ is $1$. If we replace $\sigma$ with
Identity, we can recover the original dense feedforward function.

\vspace{-15pt}
\paragraph{Sparse Factorized} The transformation matrix of the feedforward
function $f_{\theta_l}$ is denoted by $\theta_l \in \mathbb{R}^{D_{in} \times
D_{out}}$. Multiple works have explored matrix factorization techniques to
express the transformation matrix $\theta_l$ as a product of two matrices
$\theta_l = UV^T$, where $U \in \mathbb{R}^{D_{in} \times d}$, $V \in
\mathbb{R}^{D_{out} \times d}$.~\citet{khodak2020initialization,
tai2016convolutional} and~\citet{chen2021drone} have explored low-rank
factorization ($d << D_{out}$) as a form of structured sparsity to improve
training and inference efficiency, while~\citet{arora2018optimization}
and~\citet{guo2020expandnets} have explored overparameterized factorizations for
better generalization and faster convergence. In contrast, we use factorization
to augment the representational capacity without decreasing or increasing the
FLOPs. More precisely, let $\theta_l^{sf} \in \{ U_l, V_l \}$ denote the
parameters of this transformation, where $U_l \in \mathbb{R}^{D_{in} \times
d_{sf}}$, $V_l \in \mathbb{R}^{d_{sf} \times D_{out}}$ are sparse matrices with
$s$ fraction of their weights being sparse. The functional transformation in
this case is $\psi^{sf}_l = V_l^T\sigma(U_l^Tz_l)$. The computational FLOPs of
this transformation is $d_{sf}{\cdot}B{\cdot}(D_{in} + D_{out}){\cdot}(1-s)$.
Setting these FLOPs equal to FLOPs of $f_{\theta_l}$, we obtain $d_{sf} =
\frac{D_{in}{\cdot}D_{out}}{(D_{in} + D_{out}){\cdot}(1-s)}$. Note, setting
sparsity $s=0$, we recover a non-linear low-rank factorization with dense
matrices.
\begin{table}
    \caption{Cardinality of search space for sparsity masks of different members
    of the Sparse-IFT family.}
    \vspace{-5pt}
    \begin{center}
    \begin{small}
    \begin{sc}
    \resizebox{0.65\linewidth}{!}{
    \begin{tabular}{cc}
        \toprule
        Transformation & \begin{tabular}[c]{@{}c@{}}Cardinality of\\ Search
        Space\end{tabular} \\ \midrule Sparse Wide &
        $(k_{sw})^2{\cdot}(D_{in}{\cdot}D_{out})$ \\
        Sparse Parallel    &  $k_{sp}{\cdot}(D_{in}{\cdot}D_{out})$    \\
        Sparse Factorized  & $d_{sf}{\cdot}(D_{in} + D_{out})$   \\
        Sparse Doped       & $D_{in}{\cdot}D_{out}$            \\
        \bottomrule
    \end{tabular}
    }
    \end{sc}
    \end{small}
    \end{center}
    \label{tab:nature_of_sift_transformations}
    \vspace{-10pt}
\end{table}
\vspace{-10pt}
\paragraph{Sparse Doped} is inspired by previous works which approximate a dense
matrix with a combination of low-rank factorization and sparse
matrix~\citep{chen2021scatterbrain, thakker2021doping, udell2019big,
candes2011robust}. In our approach, we replace the feedforward function with
low-rank factorization (with rank $d_{sd}$) and an unstructured sparse weight
matrix (with sparsity $s$). Let $U_l \in \mathbb{R}^{D_{in} \times d_{sd}}, V_l
\in \mathbb{R}^{d_{sd} \times D_{out}}$ denote the low-rank matrices, and
$\theta_l^{sd} \in \mathbb{R}^{D_{in} \times D_{out}}$ denote the matrix with
unstructured sparsity. The functional transformation, in this case, is given by
$\psi^{sd}_l = V_l^T(U_l^Tz_l) + \sigma((\theta_l^{sd})^Tz_l)$. The
computational FLOPs associated with this transformation are
$B{\cdot}d_{sd}{\cdot}(D_{in} + D_{out}) +
(1-s){\cdot}B{\cdot}D_{in}{\cdot}D_{out}$. Setting these FLOPs equal to FLOPs of
$f_{\theta_l}$, we obtain $d_{sd} = \frac{s{\cdot}D_{in}{\cdot}D_{out}}{(D_{in}
+ D_{out})}$. Note, as $s \to 0$ and $d_{sd} \to 0$, the low-rank component of
disappears, and we can recover the dense feedforward function as a special case
by setting $\sigma$ to Identity.

\vspace{-10pt}
\paragraph{Cardinality of Search Space}
\label{subsec:cardinality} 
Increasing the sparsity mask search space with Sparse-IFT is anticipated to
enhance training efficiency, as indicated by prior works~\citep{ramanujan2020s,
liu2022unreasonable, stosic2021search}. The likelihood of finding a lottery
ticket in a randomly initialized network increases with network
width~\citep{ramanujan2020s}. Both \citet{liu2022the} and
\citet{stosic2021search} show that expanding the search space through increased
width or depth improves accuracy. The search space cardinality, defined as the
weights a sparse training method can explore, is detailed in
Table~\ref{tab:nature_of_sift_transformations}. Sparse Wide, Sparse Parallel,
and Sparse Factorized scale with width, parallel branches, and hidden dimension
size, respectively. Sparse Doped maintains a constant search space by allocating
FLOPs between a low-rank and an unstructured sparse weight matrix. Therefore,
DST becomes crucial for effectively traversing this larger parameter subspace,
as discussed in Section~\ref{sec:dst}.
\vspace{-5pt}
\section{Sparse-IFT Ablation Studies} 

\begin{figure*}[]
    \vspace{-10pt}
    \centering
    \subfigure{\includegraphics[width=0.33\textwidth]{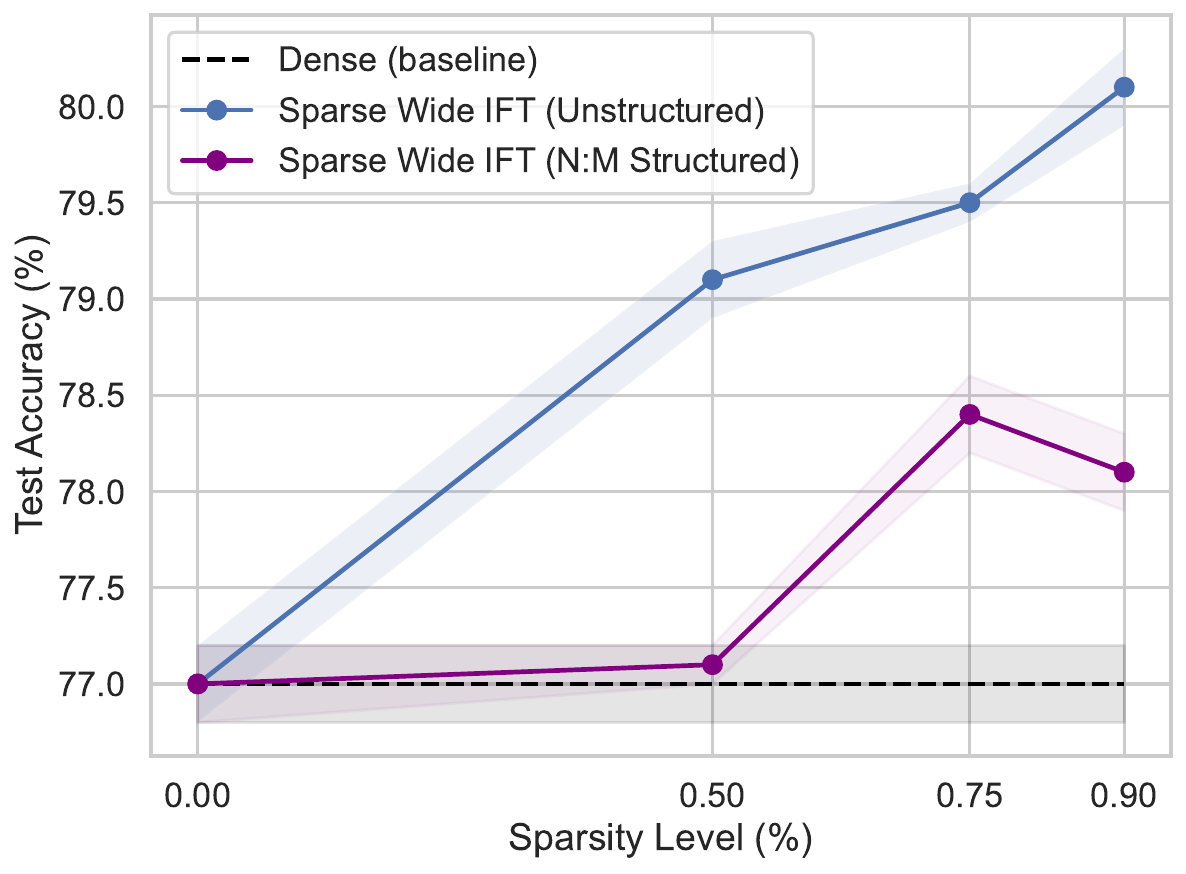}}
    \subfigure{\includegraphics[width=0.33\textwidth]{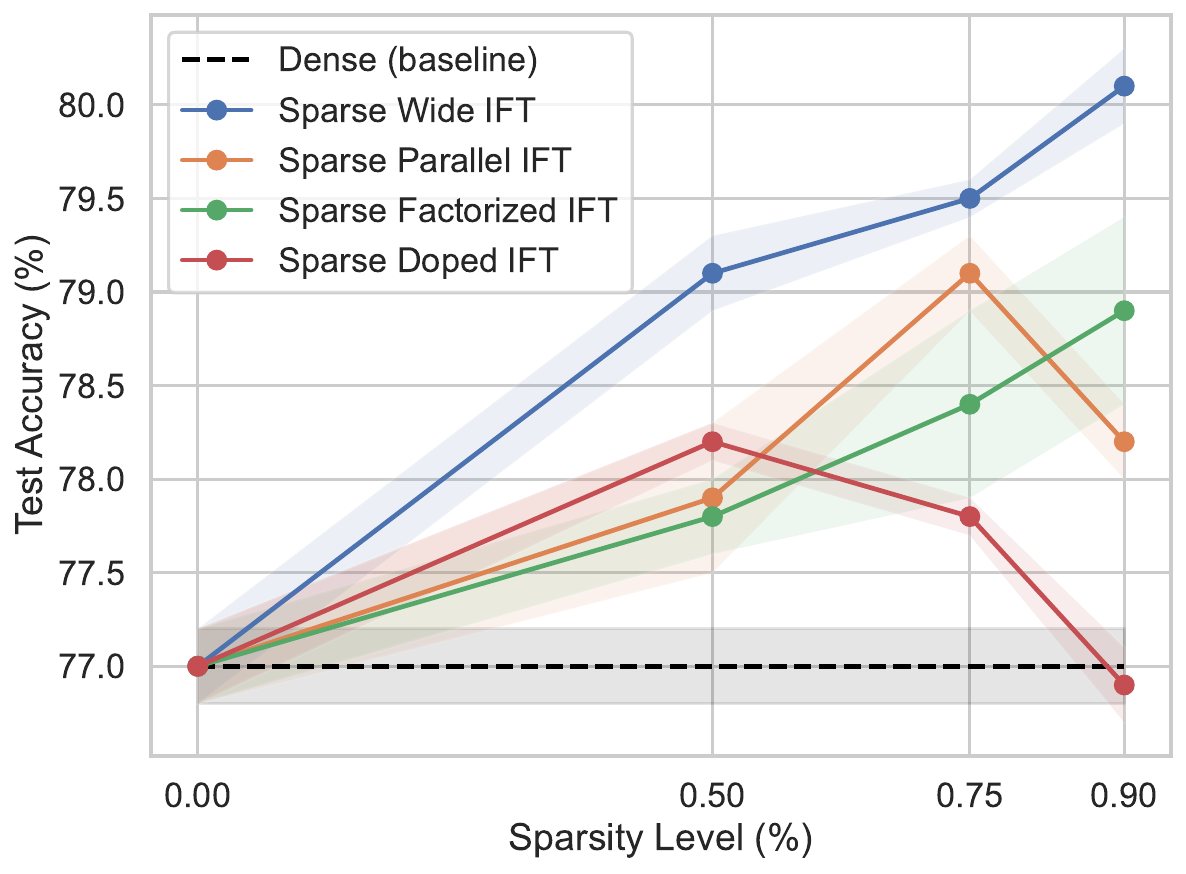}}
    \subfigure{\includegraphics[width=0.33\textwidth]{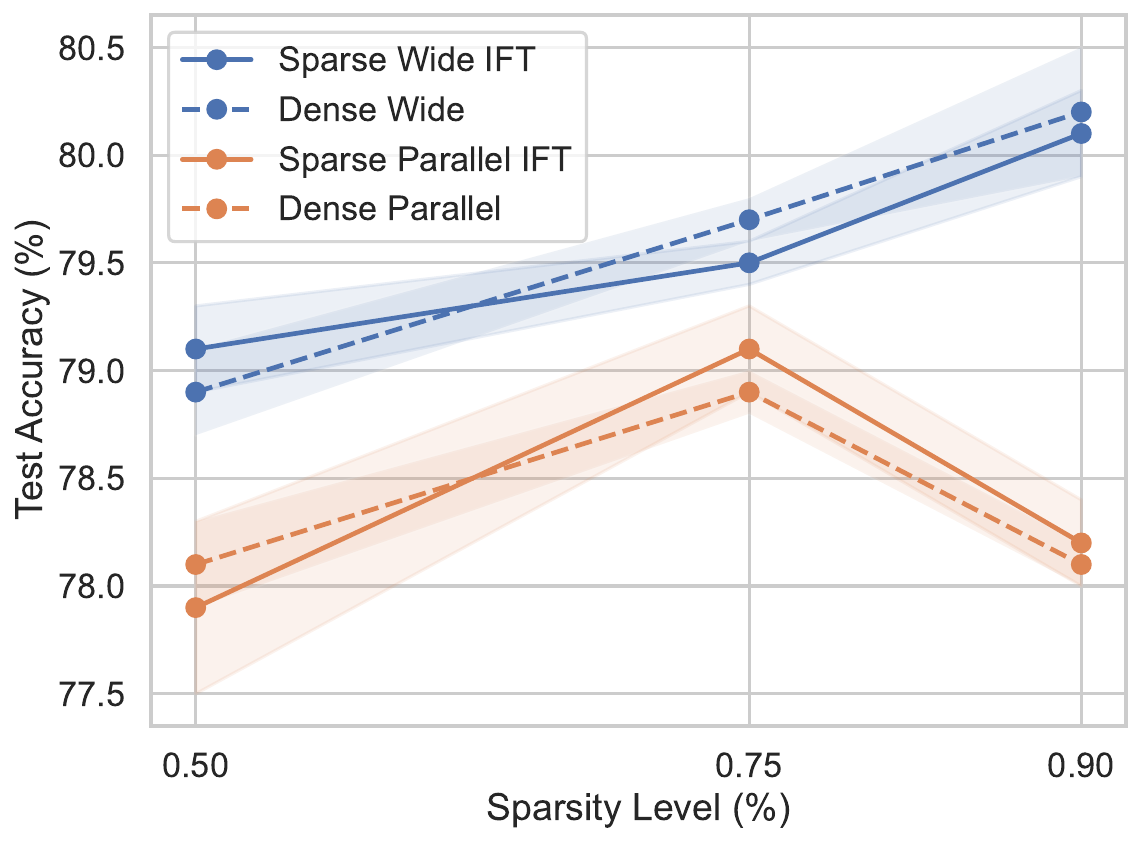}}
    \vspace{-20pt}
    \caption{Ablation studies with Sparse-IFT on the ResNet-18 model for
    CIFAR-100 across sparsity $\in \{50\%, 75\%, 90\%\}$. (left) Sparse Wide IFT
    trained with dynamic unstructured and structured sparsity. (middle)
    Sparse-IFT family members trained with RigL, where Sparse Wide performs the
    best. (right) Sparse Wide IFT trained in a sparse and dense manner.}
    \label{fig:cifar100_ablations}
    \vspace{-10pt}
\end{figure*}

In this section, we present a comprehensive analysis of Sparse-IFT networks,
focusing on their training methodologies and design considerations. First, we
compare static sparse training with DST, highlighting DST's superior performance
in handling larger parameter spaces through empirical results using the
ResNet-18 architecture on CIFAR-100. Then, we explore critical design aspects of
Sparse-IFT, including the role of non-linearities, and the benefits of dynamic
unstructured sparsity over structured sparsity. Finally, we evaluate the
efficacy of DST by comparing it against densely trained Sparse-IFT models.
\vspace{-5pt}
\subsection{Impact of Sparse Training Techniques}
\label{sec:dst}
\begin{table}
    \caption{Sparse Wide IFT with ResNet-18 trained using various sparse
training methods on CIFAR-100 across different sparsity levels (columns). Best
accuracy for each sparse training method is highlighted in bold. } 
    \vspace{-2pt}
    \centering
    \begin{small}
    \begin{sc}
    \resizebox{\linewidth}{!}{
        \begin{tabular}{ccccc}
            \hline
            \multicolumn{1}{c|}{Dense}                           & Sparse Method
            & 0.50                                     & 0.75           & 0.90
            \\ \hline
            \multicolumn{1}{c|}{\multirow{7}{*}{77.0 $\pm$ 0.2}} & Static &
            \textbf{78.5 $\pm$ 0.3} & 78.3 $\pm$ 0.1 & 78.2 $\pm$ 0.3 \\
            \multicolumn{1}{c|}{}                                & SNIP &
            \textbf{77.8 $\pm$ 0.3}                  & 77.0 $\pm$ 0.2 & 75.8
            $\pm$ 0.2                           \\
            \multicolumn{1}{c|}{}                                & GraSP &
            \textbf{77.7 $\pm$ 0.3}                  & 76.5 $\pm$ 0.3 & 76.5
            $\pm$ 0.3                           \\
            \multicolumn{1}{c|}{}                                & FORCE &
            \textbf{77.2 $\pm$ 0.3}                  & 76.9 $\pm$ 0.3 & 75.4
            $\pm$ 0.4                           \\
            \multicolumn{1}{c|}{}                                & SET & 78.8
            $\pm$ 0.1                           & 79.2 $\pm$ 0.2 & \textbf{79.8
            $\pm$ 0.2} \\
            \multicolumn{1}{c|}{}                                & RigL & 79.1
            $\pm$ 0.2                           & 79.5 $\pm$ 0.1 & \textbf{80.1
            $\pm$ 0.2} \\  
            \multicolumn{1}{c|}{}                                 & GraNet &
            79.2 $\pm$ 0.2                           & 79.6 $\pm$ 0.2 &
            \textbf{80.0 $\pm$ 0.2}  \\ \bottomrule
            \end{tabular}
    }
    \end{sc}
    \end{small}
\label{tab:dynamic_sparsity_importance}
\vspace{-15pt}
\end{table}

This section provides a comparative analysis of Sparse-IFT networks trained with
two classes of methods: static sparse training and DST. The focus is on
demonstrating DST's effectiveness in navigating larger parameter spaces, as
evidenced by previous research~\citep{huang2023dynamic,tai2022spartan}. Our
empirical results consistently show DST's superiority over static sparse
training. All experiments utilize the ResNet-18 architecture on CIFAR-100 with
published settings~\citep{devries2017improved}. Detailed model information and
hyperparameters are available in Appendix \ref{app:pt-implement}, and all
results are averaged over 3 seeds.

Sparse-IFTs employ unstructured sparsity in its transformations. This study
investigates the impact of sparse training methods on various Sparse-IFT
configurations, focusing on Sparse Wide IFT with sparsity $\in \{50\%, 75\%,
90\%\}$. In Table~\ref{tab:dynamic_sparsity_importance}, we evaluate: random
static sparsity, SNIP~\citep{lee2018snip}, GraSP~\citep{wang2020picking},
FORCE~\citep{de2020progressive}, SET~\cite{mocanu2018},
RigL~\cite{evci2020rigging} and GraNet~\citep{liu2021sparse}. SET, RigL, and
GraNet are DST methods, with SET updating the mask randomly, RigL updating it
with gradient information and GraNet incorporating gradual magnitude
pruning~\citep{Zhu2017ToPO} with RigL. Pruning at Initialization (PaI) methods
(e.g., SNIP, GraSP, FORCE) and GraNet increase training FLOPs due to non-uniform
sparsity and dense-to-sparse training. We address this by adjusting target
sparsity levels to align Sparse-IFT training FLOPs with the dense baseline (see
Appendix~\ref{app:control_flops}). In Iso-FLOP scenarios, PaI methods
underperform because they heavily prune parameter-rich layers to match target
sparsity levels, leading to layer-collapse and poor gradient flow. Furthermore,
DST methods consistently outperform static sparsity, with improvements
persisting at higher sparsity levels. Sparse-IFTs expand the sparse mask-weight
space $\propto$ sparsity, benefiting DST in thorough exploration and
exploitation within this space. While RigL and GraNet attain similar
performance, RigL is chosen as the sparse training method for simplicity in all
experiments.
\vspace{-15pt}
\subsection{Assessing the Effects of Architecture Variations}
\label{sec:sift_ablations}
This section analyzes different design considerations for Sparse-IFT by first,
exploring the role of non-linearities in enhancing representational capacity.
Then, the advantage of training with dynamic unstructured sparsity over
structured is investigated. Next, we compare between densely and sparsely
trained Sparse-IFT models. Finally, by applying top-performing Sparse-IFTs to
efficient vision models, these insights contribute to a synthesized framework.
\vspace{-15pt}
\paragraph{Importance of Using Non-Linear Activations}
For some of the Sparse-IFT members, we draw inspiration from linear
overparameterization methods, which fold the feedforward function into a dense
matrix post-training~\cite{ding2021repvgg, ding2021diverse, guo2020expandnets,
ding2019acnet}. Our method enhances representational capacity through an
Iso-FLOP transformation without increasing training FLOPs. Maintaining original
dense FLOP levels eliminates the need for post-training modifications, enabling
efficient inference and incorporation of non-linearities (i.e., ReLU) in
Sparse-IFT. Experiments on ResNet-18 on CIFAR-100 show notable accuracy gains
across all sparsity levels with non-linear activations. For example, at 90\%
sparsity, using non-linearities in Sparse Factorized IFT yields a 1.8\% accuracy
increase over the dense baseline, in contrast to a 0.5\% decrease without
non-linearities. These findings extend to all Sparse-IFT members (see
Appendix~\ref{app:nonlinear-importance} for details). The accuracy improvements
at all sparsity levels highlight the effectiveness of incorporating non-linear
activations in Sparse-IFT.
\vspace{-15pt}
\paragraph{Unstructured vs. Structured Sparsity}
We compare dynamic unstructured and structured sparsity using Sparse-IFT.
Unstructured sparsity explores all mask variations, but most hardware
accelerators do not support unstructured sparse acceleration. Prior works have
investigated structured sparsity, such as low-rank and block-sparse matrices,
for wall-clock speed-ups~\cite{khodak2020initialization, chen2021drone,
hubara2021accel, dao2022monarch}. We explore structured sparsity through
Iso-FLOP configurations with Sparse Wide IFT, employing low-rank factorization
and N:M sparsity for GPU acceleration. In Figure~\ref{fig:cifar100_ablations}
(left plot), we compare dynamic unstructured sparsity with N:M transposable
structured sparsity~\citep{hubara2021accel} using Sparse-IFT. The latter
demonstrates improvements over the dense baseline at 75\% and 90\% sparsity
levels. Results also indicate that N:M block sparsity outperforms low-rank
factorization (see Appendix~\ref{app:structured_sparsity}). However,
unstructured sparsity still gives the highest gains, as N:M sparsity has reduced
mask diversity in block-sparse matrices~\citep{hubara2021accel},  therefore, we
adopt unstructured sparsity in all subsequent experiments.

\vspace{-15pt}
\paragraph{Sparse-IFT ResNet-18} 
We assess all Sparse-IFT family members with ResNet-18 on CIFAR-100 across
different sparsity levels. The middle plot of
Figure~\ref{fig:cifar100_ablations}, highlights the best accuracy achieved by
each Sparse-IFT member. All members exhibit substantial accuracy improvements
compared to the dense baseline (77\%), using the same FLOPs. Sparse Wide
consistently performs the best, while Sparse Doped is the only member not
gaining accuracy at higher sparsity. This is attributed to Sparse Doped
maintaining constant search space by distributing FLOPs between low-rank and
unstructured sparse matrices (see
Table~\ref{tab:nature_of_sift_transformations}), leading to a decrease in active
weights in the unstructured matrix. In
Appendix~\ref{app:sift_vs_extendedsparse}, we compare Sparse-IFT against other
DST baselines under the same training efficiency setup by extending the training
steps, showing Sparse-IFT outperforms them significantly at $s \in \{50\%, 75\%,
90\%\}$. Since, Sparse Parallel and Sparse Wide perform the best across
ablations, we use these two IFTs for the main experiments.
\vspace{-15pt}
\paragraph{Sparse-IFT vs. Dense Overparametrization}
A crucial element in the success of Sparse-IFT lies in its efficient exploration
of the search space. In this section, to benchmark this exploration, we
establish an upper bound by training the Sparse-IFT architectures in a dense
manner (with sparsity levels $s \in \text{50\%, 75\%, 90\%}$). In
Figure~\ref{fig:cifar100_ablations}, the right plot compares the sparse and
dense versions of Sparse Wide and Sparse Parallel IFTs. Both Sparse-IFT members
excel in exploring a large search space with DST, achieving accuracy comparable
to their dense counterparts without the computational overhead. These results
highlight that the sparsity search in DST approaches optimality and can achieve
accuracy comparable to that of densely trained models. This efficiency does not
compromise accuracy and offers substantial computational benefits, especially on
hardware optimized for sparsity (further discussed in Section 6).
\vspace{-5pt}
\begin{table}
    \caption{Sparse Wide IFT with various efficient architectures on CIFAR-100
across different levels of sparsity (columns).} 
    \vspace{-5pt}
    \centering
    \begin{sc}
    \begin{small}
    \resizebox{0.85\linewidth}{!}{
    \begin{tabular}{c|ccc}
        \toprule
        	    & Dense             & 0.50 & 0.75 \\ \midrule MobileNetV2 & 72.4
        $\pm$ 0.2    & 73.4 $\pm$ 0.2 & \textbf{73.7 $\pm$ 0.2}  \\
        MobileViT-S & 73.5 $\pm$ 0.1    & 74.6 $\pm$ 0.2 & \textbf{74.8 $\pm$
        0.2} \\
        BotNet-50   & 79.8 $\pm$ 0.2     & 80.3 $\pm$ 0.3 & \textbf{80.9 $\pm$
        0.3} \\
        \bottomrule
    \end{tabular}
    }
    \end{small}
    \end{sc}
\label{tab:mbv2-cifar}
\vspace{-5pt}
\end{table}
\vspace{-20pt}
\paragraph{Efficient Architectures}
\label{subsec:efficient_archs}
To assess Sparse-IFT's robustness across diverse set of models, we evaluate it
on architectures optimized for efficient inference
(MobileNetV2~\citep{sandler2018mobilenetv2} and
MobileViT~\citep{mehta2021mobilevit}) and efficient training
(BotNet~\citep{srinivas2021bottleneck}). Applying Sparse Wide IFT to dense
layers significantly improves test accuracy across all architectures (refer to
Table~\ref{tab:mbv2-cifar}). Similarly, utilizing the Sparse Parallel IFT
consistently enhances performance across all architectures (see
Appendix~\ref{app:efficientcv}). We evaluate the best-performing model,
BotNet-50, on ImageNet, where the Sparse-IFT variant outperforms dense by 1\%
(see Section~\ref{subsec:imagenet}). We provide additional experimental setup
details in Appendix~\ref{app:pt-implement}. In summary, Sparse-IFT significantly
improves test accuracy across all efficient architectures, demonstrating its
robustness and effectiveness.
\vspace{-15pt}
\section{Spectral Analysis of DST in Sparse-IFT}
In this study, we investigate the intricate properties of Sparse-IFT networks
and their training dynamics. We analyze the benefits of Sparse-IFT networks
trained with DST by analyzing the Ramanujan Gap and Spectral Gap
characteristics. Ramanujan graph structures which are known to exhibit sparsity
and high connectivity like expander graphs, are investigated to reveal their
correlation with the final performance of sparse networks. Our analysis
evaluates the impact of model parameters and graph connectivity on the
effectiveness of DNNs with Sparse-IFTs, aiming to provide insights into the
training dynamics of Sparse-IFT models. Inspired by~\citet{hoang2023revisiting,
hoang2023dont}, in this analysis, we interpret the ResNet-18 model as a series
of bipartite compute graphs, where each layer, $\{\theta_1,\ldots, \theta_L\}$
in an $L$ layered sparse DNN, takes the form of a square adjacency matrix $A$.
\citet{hoang2023revisiting} proposed several graph metrics inspired by Ramanujan
properties for characterizing sparse networks, via: 1) \textbf{Ramanujan Gap}:
$\Delta r = 2 \ast \sqrt{d-1} - \hat{\mu}(A)$, and $\Delta r_{imdb} =
\frac{1}{\abs{K}} \sum_{i=1}^{\abs{K}}(2\sqrt{d_i - 1} - \hat{\mu}(A_{K_i}))$,
where $d$ is the average edge per node, and $\hat{\mu}(A)$ is the non-trivial
eigenvalue of $A$. Here, $\Delta r$ is the conventional view of measuring gap
between Ramanujan's upper bound $2 \ast \sqrt{d-1}$ and $\hat{\mu}(A)$. $\Delta
r$ measures the network's degree of connectivity to reveal the flow of
information propagation. $\Delta r_{imdb}$~\citep{hoang2023revisiting}, the
Iterative Mean Difference Bound (imdb), evaluates the average connectivity
boundary across all subgraphs $K$ within $A$. A higher $\Delta r$ in sparse networks signifies efficient information flow,
gradient propagation, and a well-separated spectrum in the adjacency matrix of
sparse weights; indicating robust and efficient representation. In addition, an
increasing $\Delta r_{imdb}$ indicates more extensive connectivity boundaries
within subgraphs, enhancing communication among nodes and promoting stronger
connections. 2)~\textbf{Weighted Spectral Gap}: $\lambda = \mu_0(\abs{\vb*{W}})
- \hat{\mu}(\abs{\vb*{W}})$, and $\lambda_{imsg} =
\frac{1}{\abs{K}}\sum_{i=1}^{\abs{K}}(\mu_{0}(\abs{\vb*{W}_{K_i}}) -
\hat{\mu}(\abs{\vb*{W}_{K_i}}))$. Here, the gap between $\mu_0$, the trivial
parameters, and $\hat{\mu}$, the non-trivial eigenvalues of $\vb*{W}$, the
weighted adjacency matrix, is denoted as $\lambda$, the weighted spectral gap.
Then, $\lambda_{imsg}$~\citep{hoang2023dont} is the iterative version which
takes into account all subgraphs $K$ within $\vb*{W}$. A higher $\lambda_{imsg}$
indicates enhanced spectral separation between $\mu_0$ and $\hat{\mu}$ of
$\vb*{W}$, implying a more distinct and well-defined spectral structure within
subgraphs. This improved separation in the spectrum, represented by a higher
$\lambda$, facilitates better isolation of meaningful signals. We train Sparse
Wide and Sparse Parallel ResNet-18 models at 50\% sparsity on CIFAR-100. Then,
we generated a Pareto curvature heatmap, considering weight magnitudes and graph
topological structure details (see Figure~\ref{fig:graph_analysis}). See
Appendix~\ref{app:graph_analysis_sift_dst} for a detailed analysis. 
\begin{figure}
    \vspace{-0.1in}
    \centering
    \subfigure{\includegraphics[width=0.23\textwidth]{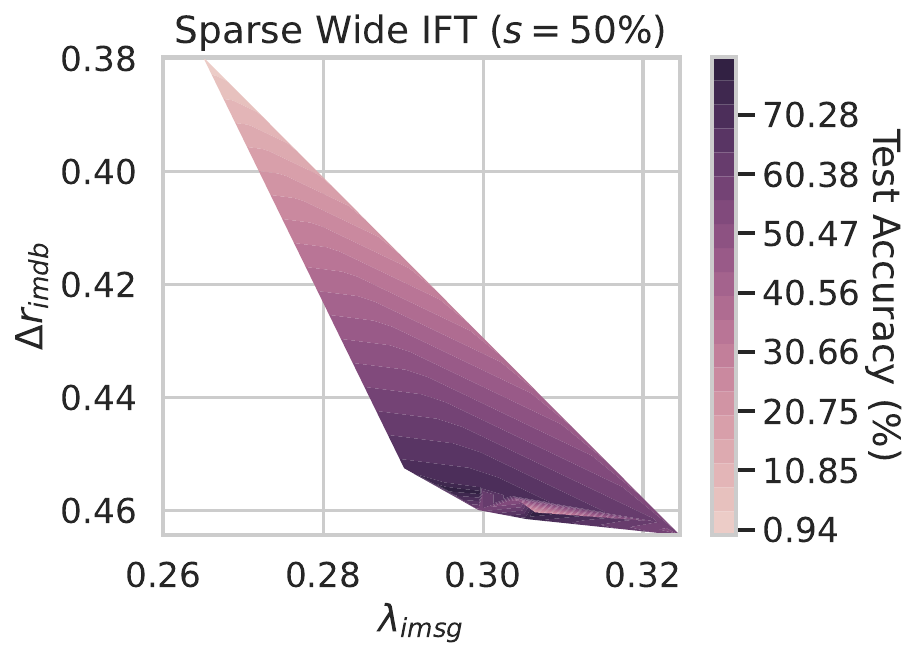}}
    \subfigure{\includegraphics[width=0.23\textwidth]{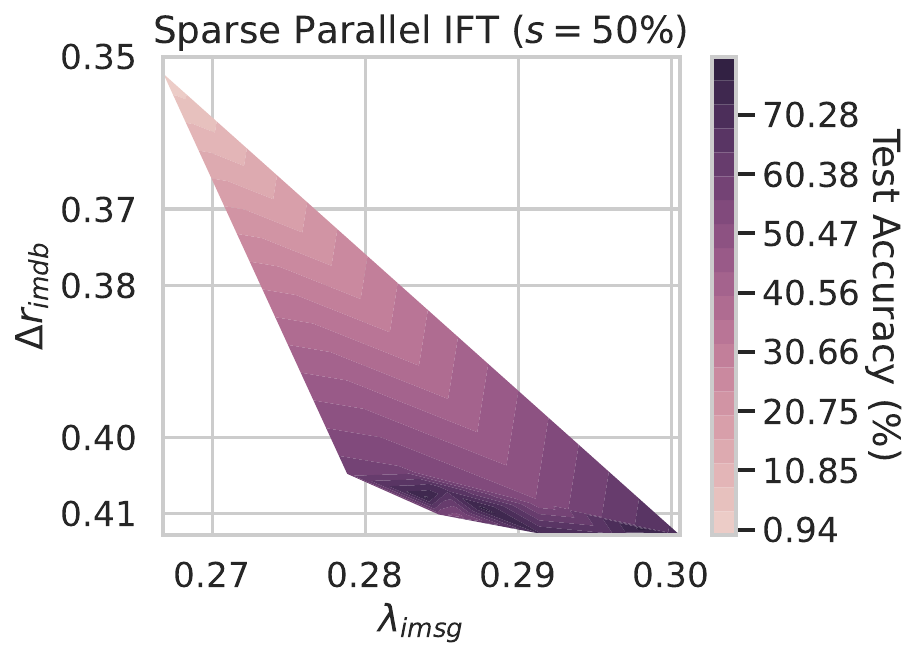}}
    \\ \vspace{-10pt}
    \subfigure{\includegraphics[width=0.23\textwidth]{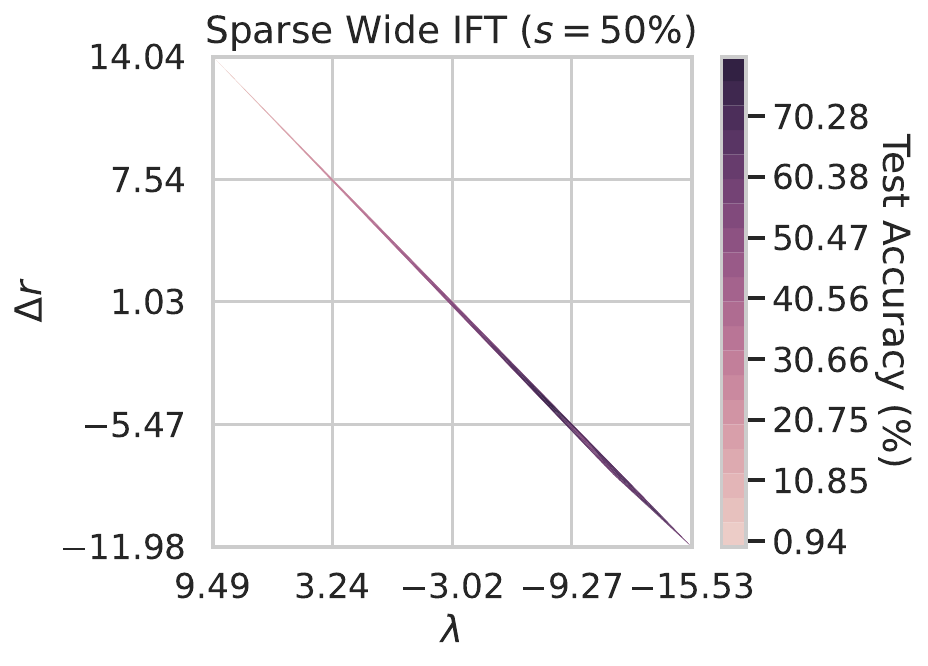}}
    \subfigure{\includegraphics[width=0.233\textwidth]{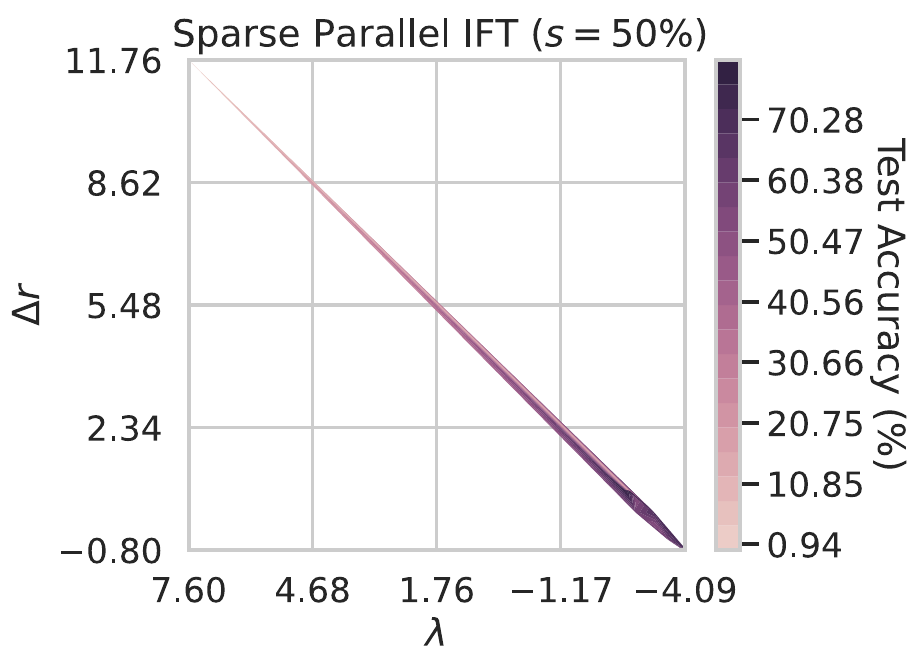}}
    \vspace{-15pt}
    \caption{The relationship between the structure and weights of Sparse-IFT
    ResNet-18 networks are analyzed through a graph perspective in terms of
    performance. Top row: we assess the relationship between $\Delta r_{imdb}$
    and $\lambda_{imsg}$. Bottom row: investigates the correlation between
    $\Delta r$ and $\lambda$. The Pareto curvature heatmap visually represents
    the classification performance, with varying color gradients symbolizing the
    spectrum from low to high test accuracy on CIFAR-100.}
    \vspace{-5pt}
    \label{fig:graph_analysis}
    \vspace{-10pt}
\end{figure}

\textbf{$\vb*{\Delta r_{imdb}}$ and $\vb*{\lambda_{imsg}}$ Analysis:} In the
initial to middle stages of training, RigL's dynamic pruning and regrowth
increases $\Delta r_{imdb}$ for the network to \textit{explore} diverse
connectivity patterns (see top row, Figure~\ref{fig:graph_analysis}). Subsequent
pruning removes less critical connections, diversifying subgraphs in the
adjacency matrix $A$. Later stages witness a decrease in $\Delta r_{imdb}$ as
the network converges to more focused and organized connectivity patterns. RigL
prioritizes crucial connections, \textit{exploiting} an efficient subgraph
structure linked with highly accuracy regions of Sparse-IFT models. Early in
training, increasing $\lambda_{imsg}$ suggests successful isolation of different
modes. Pruning leads to a distinct separation between dominant and less dominant
modes. The subsequent $\lambda_{imsg}$ decrease signals the network's
convergence to a more specialized representation, emphasizing key spectral
components and diminishing the influence of less critical modes. While both
Sparse Wide and Sparse Parallel IFTs show increasing $\Delta r_{imdb}$, the
larger search space cardinality in Sparse Wide facilitates the emergence of
diverse subgraph structures within each layer, allowing for a richer set of
connections between nodes; resulting in a higher maximum $\Delta r_{imdb}$.
Similarly, Sparse Wide has a higher maximum $\lambda_{imsg}$ compared to Sparse
Parallel, indicating the emergence of subgraphs with more distinct spectral
properties. \\
\textbf{$\vb*{\Delta r}$ and $\vb*{\lambda}$ Analysis:}
Figure~\ref{fig:graph_analysis}'s bottom row reveals a strong correlation
between $\Delta r$ and $\lambda$. $\Delta r$ initially decreases, indicating a
temporary relaxation of spectral constraints during dynamic pruning with RigL.
Subsequently, it maximizes in the final training stages, signifying RigL's
ability to guide the network to reorganize its connectivity, promoting more
structured and favorable spectral characteristics. Similarly, $\lambda$ follows
a pattern of initial decrease and later maximization. This implies that RigL's
dynamic sparsity initially results in less optimal weight organization
concerning spectral properties. However, RigL's iterative pruning and rewiring
dynamically adapts the network, aligning weights to enhance spectral
characteristics and increase the spectral gap.  Our analysis demonstrates that
DST, as exemplified by RigL, outperforms static sparse training by optimizing
spectral characteristics for Sparse-IFT; facilitating improved connectivity
patterns and a more favorable spectral profile.
\vspace{-5pt}
\section{Empirical Evaluation}\label{sec:empirical_res} Building on insights
gained from our ablations discussed in Section~\ref{sec:sift_ablations}, we
apply Sparse-IFTs to ImageNet, also demonstrating its advantages for transfer
learning in various computer vision tasks. Additionally, we highlight the
benefits of Sparse-IFT in the domain of NLP by presenting results on
pre-training GPT~\citep{brown2020language}.
\vspace{-5pt}
\subsection{ImageNet and Transfer Learning}
\label{subsec:imagenet}
We apply the best-performing Sparse-IFT transformations (Sparse Wide IFT and
Sparse Parallel IFT) from CIFAR-100 to ImageNet using ResNet-18. We follow
published training settings for ImageNet~\citep{nvidia2023gpuperf}. Both
Sparse-IFT families achieve significantly higher accuracy compared to the dense
baseline (see Table~\ref{tab:resnet-i1k}). Specifically, Sparse Wide IFT
ResNet-18 at 90\% sparsity improves over the dense baseline by 3.5\% and matches
the accuracy of a dense ResNet-34 with 2× fewer training FLOPs (refer to
Figure~\ref{fig:sift_resnet_improvement}). We also apply the best-performing
transformation (Sparse Wide IFT) to ResNet-34 and BotNet-50. Increasing sparsity
consistently improves accuracy, indicating enhanced training efficiency at
higher sparsities. On BotNet-50, a hybrid ViT model, there is a 1.1\%
improvement at 90\% sparsity.
\vspace{-10pt}
\paragraph{Transfer Learning on Downstream}
\label{subsec:transfer_learning}
To show the effectiveness of pre-training our Sparse-IFT classification
backbones, we evaluate them on 1) object detection on MS COCO
2017~\cite{lin2014microsoft}, and 2) semantic segmentation on
CityScapes~\cite{cordts2016cityscapes}. For object detection, we adopt
RetinaNet~\cite{lin2017focal} from the MMDetection open-source
toolbox~\cite{mmdetection} and report results in the standardized training
setting. For semantic segmentation, we utilize DeepLabV3+~\cite{chen2018encoder}
in the MMSegmenation open-source toolbox~\cite{mmseg2020}. We evaluate ResNet-18
with Sparse Wide IFT and to ensure FLOP-equivalent comparisons with the dense
backbone, the Sparse-IFT backbones remain sparse during fine-tuning.
Appendix~\ref{app:eval_downstream} provides more details on the training setup.
We summarize our findings in Table \ref{tab:down_stream}, where using Sparse
Wide IFT ResNet-18 backbone leads to significant accuracy gains across all
metrics on both tasks.

\begin{table}
    \caption{Sparse-IFT on ImageNet. Best result for each transformation and
architecture is highlighted in bold.}
    \vspace{-5pt}
    \begin{center}
    \begin{small}
    \begin{sc}
    \resizebox{\linewidth}{!}{
    \begin{tabular}{c|c|c|ccc}
        \toprule
        Model & Dense & Transformation & 0.50 & 0.75 & 0.90 \\
\midrule
        \multirow{2}{*}{ResNet-18} &  \multirow{2}{*}{70.9 $\pm$ 0.1} & Sparse
Wide & 72.7 $\pm$ 0.1 & 73.8 $\pm$ 0.2 & \textbf{74.4 $\pm$ 0.2} \\
        &   & Sparse Parallel & 72.7 $\pm$ 0.2 & 73.2 $\pm$ 0.2 & \textbf{74.0
        $\pm$ 0.2} \\ \midrule ResNet-34 & 74.2 $\pm$ 0.1 & Sparse Wide &  75.6
        $\pm$ 0.2 & 76.4 $\pm$ 0.1 & \textbf{76.8 $\pm$ 0.3} \\ \midrule
        BotNet-50 & 77.5 $\pm$ 0.1 &  Sparse Wide & 77.9 $\pm$ 0.2 & 78.3 $\pm$
        0.2 & \textbf{78.6 $\pm$ 0.3} \\
        \bottomrule
    \end{tabular}
    }
    \end{sc}
    \end{small}
\end{center}
\vspace{-5pt}
\label{tab:resnet-i1k}
\end{table}

\begin{table}
    \caption{Sparse Wide IFT variants of ResNet-18 as backbones for: (a) object
detection on MS COCO, (b) semantic segmentation on Cityscapes.} 
    \vspace{-5pt}
    \centering
    \begin{small}
    \begin{sc}
    \resizebox{\linewidth}{!}{
    \begin{tabular}{c|c|cccc}
        \toprule
      & Metric & Dense & 0.50   & 0.75  & 0.90 \\
    \midrule
 \multirow{3}{*}{MS COCO}     &  AP      &  29.3 $\pm$ 0.1  & 31.3 $\pm$ 0.1 &
 32.8  $\pm$ 0.2  & \textbf{34.5 $\pm$ 0.2}   
 \\
      &  AP$_{50}$    &  46.2 $\pm$ 0.2  & 49.0 $\pm$ 0.2 & 51.0 $\pm$ 0.2 &
      \textbf{53.5 $\pm$ 0.2}   \\
      &  AP$_{75}$    &  30.9 $\pm$ 0.2  & 33.0 $\pm$ 0.2 & 34.8 $\pm$ 0.2 &
      \textbf{36.5 $\pm$ 0.3}   \\
    \midrule
  \multirow{2}{*}{CityScapes}     &  \text{mIoU}      &   76.7 $\pm$ 0.2   &
  77.9 $\pm$ 0.2    & 78.9 $\pm$ 0.2  & \textbf{79.1 $\pm$ 0.2}    \\
      &  \text{mAcc}      &   84.4  $\pm$ 0.2   &  85.1 $\pm$ 0.2    & 85.7
      $\pm$ 0.2  & \textbf{86.0 $\pm$ 0.2}   \\
        \bottomrule
    \end{tabular}
    }
    \end{sc}
    \end{small}
\label{tab:down_stream}
\vspace{-5pt}
\end{table}

\vspace{-5pt}
\subsection{Language Modeling}
\label{subsec:nlp_results} 
We pre-train the Sparse Wide IFT GPT-3 Small model at $s \in \{50\%, 75\%\}$
from scratch on the Pile~\citep{gao2020pile} dataset using
SET~\citep{mocanu2018}, and compare against the standard dense model. All models
were trained on the Cerebras CS-2~\citep{lie_2023} following
Chinchilla~\citep{hoffmann2022an} for obtaining loss-optimal pre-trained
baseline configurations of models. We evaluate the models on 5 tasks from the
Open LLM leaderboard~\citep{open-llm-leaderboard} (i.e.,
ARC~\citep{clark2018think}, HellaSwag~\citep{zellers2019hellaswag},
MMLU~\citep{hendrycks2021measuring}, TruthfulQA~\citep{lin2022truthfulqa} and
Winogrande~\citep{DBLP:journals/corr/abs-1907-10641}), and show that the Sparse
Wide IFT GPT-3 Small at 75\% sparsity improves the average accuracy by a
noticeable 0.9\% (see Table~\ref{tab:nlp_scratch_result}). In
Appendix~\ref{app:gpt_e2e}, we provide details on the models and
hyperparameters.
\begin{table}
    \caption{Average accuracy of Sparse Wide IFT with GPT-3 Small across ARC,
     HellaSwag, TruthfulQA, MMLU and Winogrande tasks on the Open LLM
     Leaderboard.} 
    \vspace{-5pt}
    \begin{center}
    \begin{small} 
    \begin{sc}
    \resizebox{0.85\linewidth}{!}{
    \begin{tabular}{cc|cc}
        \toprule
       	Model	     & Dense & 0.50 & 0.75 \\

	\midrule
        GPT-3 Small &   33.8 $\pm$ 0.1  &  34.1 $\pm$ 0.2      &  \textbf{34.7
        $\pm$ 0.2}  \\
        \bottomrule
    \end{tabular}
    }
    \end{sc}
    \end{small}
    \end{center}
\label{tab:nlp_scratch_result}
\vspace{-15pt}
\end{table}
\vspace{-5pt}
\section{Wall-Clock Acceleration with Sparse-IFT}
\label{sec:wall_clock}
Our studies in Section~\ref{sec:empirical_res} show noticeably improved training
efficiency (test accuracy~w.r.t training FLOPs) for Sparse-IFT models. In this
section, we aim to showcase the practicality of Sparse-IFT models, providing
unique hardware insights for accelerating DNNs with unstructured sparsity, a
perspective notably absent in most existing works. Recent developments, like
specialized software kernels and hardware (e.g.,
DeepSparse~\citep{neural_magic_2021} and Cerebras CS-2~\citep{lie2023cerebras})
indicate promising gains in realizing unstructured sparsity benefits during
training and inference~\citep{thangarasa2023spdf}. This sets the stage for
examining the impact on inference and training acceleration.

\begin{figure}[!t]
    \centering
    \vspace{-5pt}
    \subfigure{\includegraphics[width=0.23\textwidth]{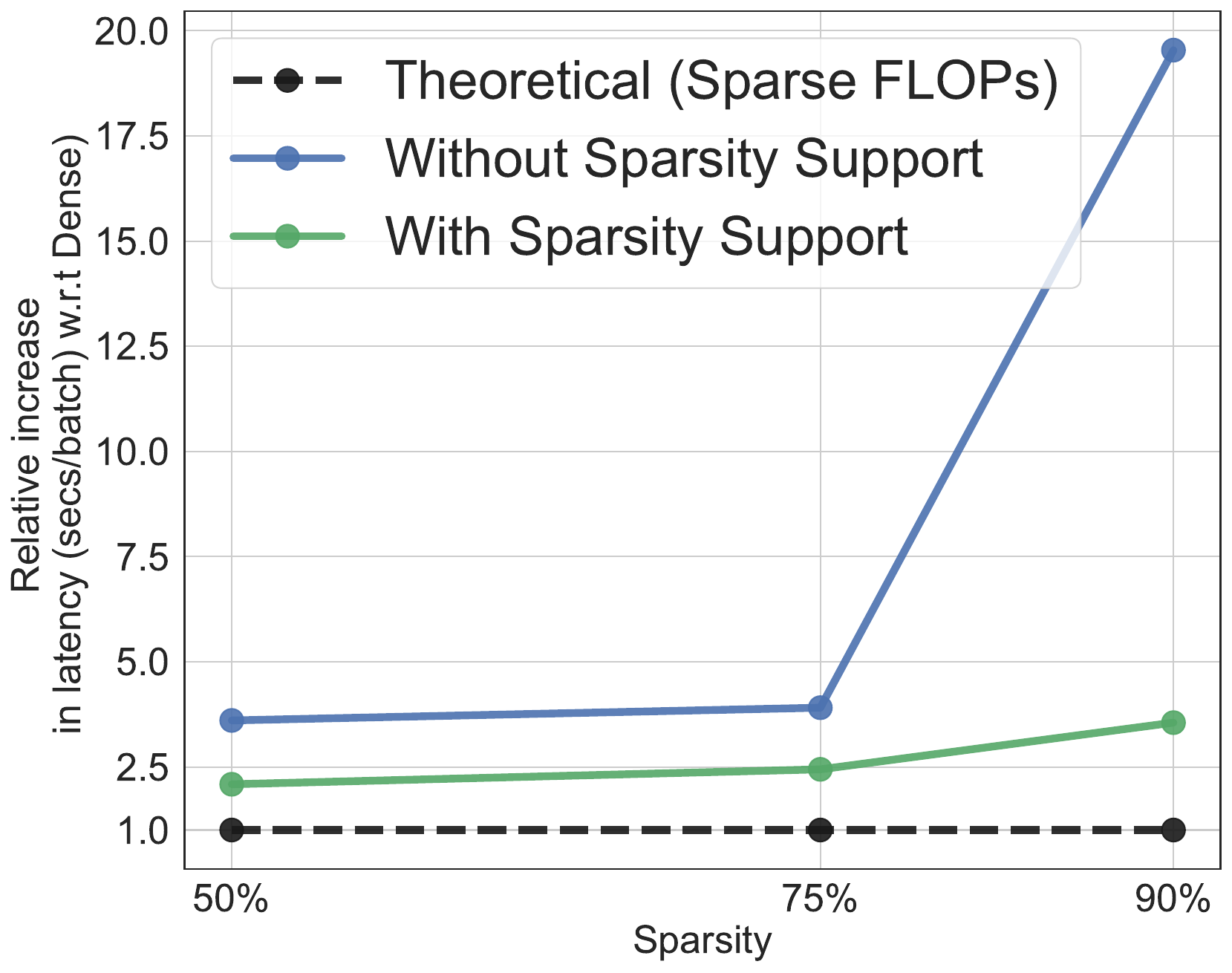}}
    \subfigure{\includegraphics[width=0.23\textwidth]{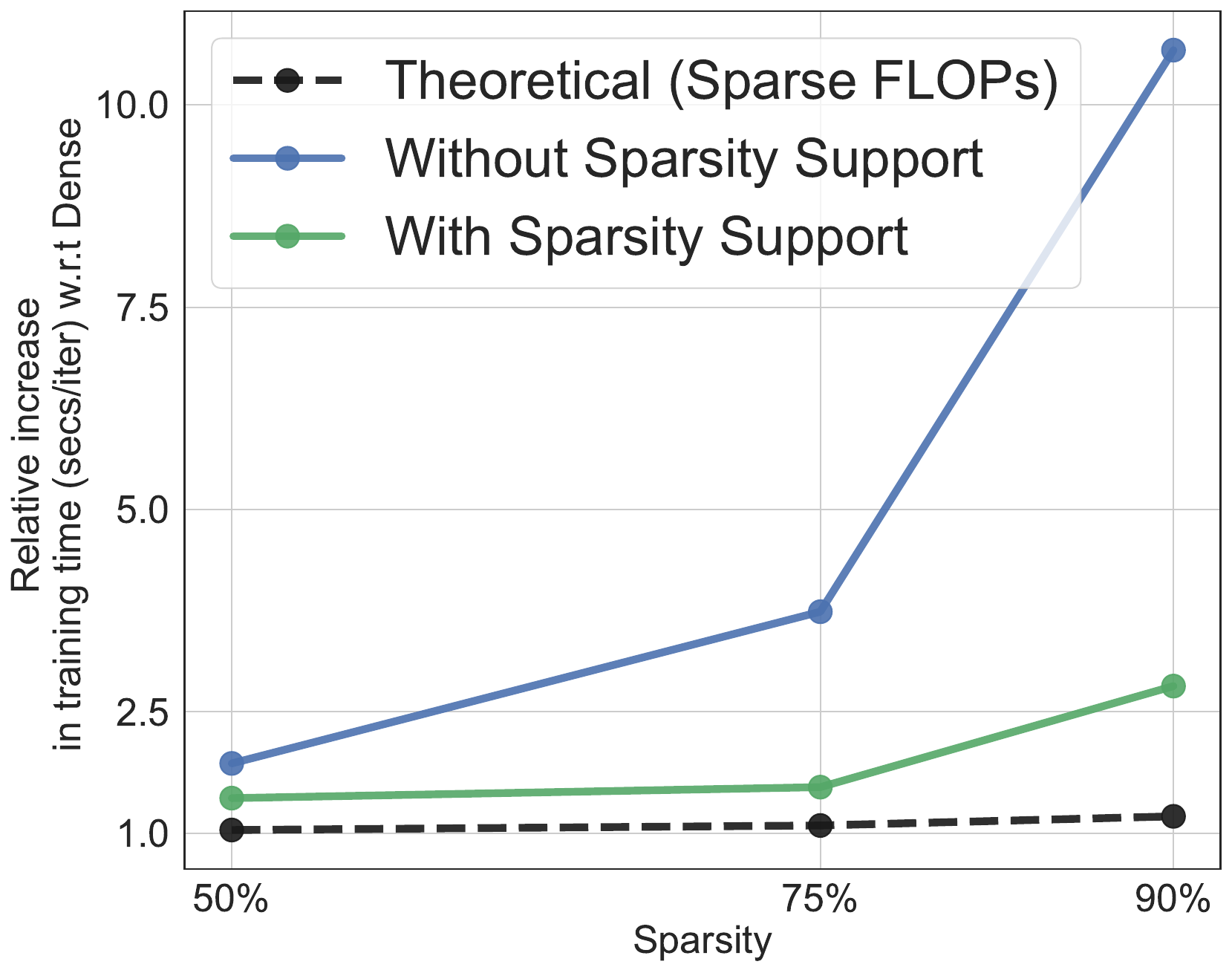}}
    \vspace{-10pt}
    \caption{Benchmarking unstructured sparsity during (left) inference on
    Neural Magic's DeepSparse runtime and (right) training acceleration on the
    Cerebras CS-2. In both setups, we measure the relative increase in latency
    or training speed for Sparse-IFT variants against the dense model.}
    \label{fig:sparsity_benchmark}
    \vspace{-10pt}
\end{figure}

\vspace{-15pt}
\paragraph{Real-World Inference Acceleration} 
We assess Sparse-IFT's inference efficiency using
DeepSparse\footnote{\href{https://github.com/neuralmagic/deepsparse}{Neural
Magic DeepSparse}}. Our setup employs a ResNet-18 model and performs batched
inference of 64 images from ImageNet at 224 × 224 resolution on Intel Cascade
Lake CPUs, known for their AVX-512 support. The latency (i.e., seconds per
batch) is compared between the dense ResNet-18 model and the Sparse Wide IFT
variants at $s \in \text{50\%, 75\%, 90\%}\}$. On an ideal hardware, FLOPs
should directly translate to wall clock time. Therefore, the inference latency
or training time for all Sparse-IFT models should match that of the dense model,
as all models are Iso-FLOP. This baseline is illustrated by the black dashed
line in the left plot of Figure~\ref{fig:sparsity_benchmark}. However, the blue
line shows the expected increases in latency on hardware \textit{without}
unstructured sparse acceleration support, like the CPUs we benchmarked on,  with
a notable 19.5x increase at $s = \text{90\%}$. In contrast, the green line
demonstrates a significant reduction in latency using DeepSparse, decreasing the
latency increase from 19.5x to 3.5x, and showing minimal overhead up to 75\%
sparsity. This emphasizes the benefits of optimized kernel support for sparse
inference acceleration, showcasing the potential for practical deployment of
Sparse-IFT models.

\paragraph{Real-World Training Acceleration} 
In the right plot of Figure~\ref{fig:sparsity_benchmark}, we evaluate the
training efficiency of Sparse-IFT on the Cerebras CS-2 system, which supports
unstructured sparse training for
LLMs\footnote{\href{https://docs.cerebras.net/en/2.1.1/wsc/how_to_guides/sparsity.html}{Cerebras
CS-2 (R2.1.1): Train a Model with Weight Sparsity}}. Our experimental setup
involves pre-training a GPT-3 model on the CS-2. We measured and compared
throughput (i.e., iterations per second), between the dense GPT-3 model and
Sparse Wide IFT variants at sparsity levels of 50\%, 75\%, and 90\%. As
previously mentioned, the theoretical baseline (black dashed line) suggests that
since both the dense model and Sparse Wide IFT configurations are Iso-FLOP,
training time should not increase with increasing sparsity. The blue line shows
the throughput for Sparse Wide IFT variants run \textit{without} any
unstructured sparse acceleration support on the Cerebras CS-2, indicating a
$\sim$10x increase in training time for the model at 90\% sparsity. A similar
degradation in performance would be expected on traditional Nvidia GPU or Google
TPU hardware as well. In contrast, the green line demonstrates the effect of
utilizing the Cerebras CS-2's unstructured sparse training support. Here, we
observe a significant reduction in relative training time, bringing down the
increase from $\sim$10x to 2.82x at 90\% sparsity. Additionally, for sparsity
levels up to 75\%, we note minimal overhead compared to the dense model.
Detailed benchmarking setups are in Appendix~\ref{app:benchmark}. 

While we do not achieve perfect FLOPs translation with Sparse-IFT models, our
promising results highlight the importance of ML software and hardware co-design
for leveraging sparsity. The interaction of layer dimensions, sparsity, and
overhead, influenced by hardware architecture, necessitates co-designed sparse
techniques for optimal performance. Our work showcases algorithmic advancements
over prior sparse methods, emphasizing the benefits of sparse training and
winning \textit{the hardware lottery}~\citep{2020shooker}.
\vspace{-10pt}

\section{Related Work}
Our work aligns with research on overparameterization and sparsity in DNN
training. The required modeling capacity for a given task is often unknown,
leading to training overparameterized models to fully exploit learning
capabilities before compressing them into smaller subnetworks.
\vspace{-10pt}
\paragraph{Overparameterization}~\citet{nakkiran2021deep} show that DNNs benefit
from overparameterization. Subsequently, several studies have capitalized on
overparameterization by scaling the size of models~\cite{rae2021scaling,
goyal2022vision} and augmenting existing DNNs to boost modeling capacity and the
accuracy of trained networks~\cite{shuxuan2020expandnet, ding2019acnet,
ding2021repvgg, jinming2022doconv, mobileone2022, liu2022more}. These methods
use linear parameterizations of the model, making them highly inefficient to
train, and are focused on improving inference throughput. In contrast, our work
is focused on improving the modeling capacity using sparse non-linear
parameterizations. Sparse-IFT enhances accuracy without increasing training and
inference FLOPs compared to the baseline dense model.
\vspace{-25pt}
\paragraph{Sparse Network Training} The Lottery Ticket
Hypothesis~\cite{frankle2018lottery, frankle2020linear} shows that accurate
sparse subnetworks exist in overparameterized dense networks but require
training a dense baseline to find. Other approaches have proposed frameworks for
identifying lottery tickets~\citep{hattie2019supermask, ma2022effective} but
still require a lot of compute resources. Following this, various attempts have
been made to find the optimal sparse subnetwork in a single shot. These methods
either try to find the subnetworks at initialization~\cite{tanaka2020pruning,
wang2020picking, de2020progressive, lee2018snip} or dynamically during
training~\cite{mocanu2018, evci2020rigging, jayakumar2020top, raihan2020sparse}.
However, given a fixed model capacity, these methods tradeoff accuracy relative
to the dense baseline to save training FLOPs. ~\citet{stosic2021search}
and~\citet{ramanujan2020s} increase the search space during sparse training to
retain accuracy; however, do not guarantee FLOPs savings. In contrast to these
methods, our work introduces a set of non-linear sparse transformations, which
increase the representational capacity of the network. This approach does not
introduce a new sparse training algorithm, but instead improves the search space
of existing methods, leading to improved generalization while being efficient to
train.

\vspace{-10pt}
\paragraph{Iso-Parameter vs. Iso-FLOP} Recent works have focused on improving
generalization at high sparsity levels. Techniques such as the
Erd\"{o}s-R\'{e}nyi-Kernel~\cite{evci2020rigging}, Ideal Gas
Quota~\cite{chen2022sparsity}, and parameter leveling~\cite{golubeva2021are}
employ layer-wise sparsity distributions in sparse training to boost accuracies.
These methods, however, target scenarios where models have a fixed parameter
budget (i.e., Iso-Parameter), which does not equate to similar training FLOPs as
the original dense model. Our work highlights that while transformer-based NLP
networks may not show significant differences between Iso-Parameter and Iso-FLOP
optimization, this distinction becomes critical in CV networks. In CNNs and
heterogeneous ViTs~\cite{pyramidvit2021, scalablevit2021, cvt2021}, the uneven
distribution of parameters and computational costs across layers necessitates a
distinct approach. Optimizing for Iso-Parameter typically involves pruning
later, parameter-rich layers, thus maintaining performance but not significantly
reducing computational costs. Conversely, optimizing for Iso-FLOP shifts pruning
to early, FLOP-intensive layers, enhancing performance by addressing both
computational demands and pruning needs. Unlike variable sparsity techniques
that adapt to different computational and memory demands across layers, our
method employs a uniform sparsity approach, ensuring consistent FLOP reductions
across all layers. This aligns computational costs closely with those of a fully
dense model, achieving significant computational efficiencies without
compromising performance.

\vspace{-10pt}
\paragraph{Sparse-IFT and Scaling Laws}
Recent advances in deep learning highlight the importance of scaling laws, which
provide a systematic framework for optimizing model performance as model size
increases. Pioneering scaling laws work such as ConvNeXt~\citep{convnext2023},
EfficientNet~\citep{pmlr-v97-tan19a}, large language
models~\citep{kaplan2020scaling}, and vision
transformers~\citep{alabdulmohsin2023getting} demonstrate that achieving optimal
performance typically involves tuning multiple training (e.g., learning rate,
batch sizes, etc.) and architectural (e.g., depth, width, resolution, etc.)
hyperparameters. This intricate balance necessitates extensive experimentation
and hyperparameter tuning. Sparse-IFT introduces a streamlined approach for
scaling DNNs, leveraging a single hyperparameter, the sparsity level, to enhance
model efficiency and accuracy. This method simplifies the optimization process
by eliminating the need to tune multiple factors concurrently. Future research
will explore the integration of Sparse-IFT with scaling laws to address the
challenges of scaling large models. This involves examining the interplay
between Sparse-IFT and various architectural elements, such as depth and width,
while maintaining a constant computational FLOP budget.
\vspace{-10pt} 

\section{Conclusion}
We introduced Sparse-IFT as a drop-in replacement for dense layers in DNNs,
enhancing test accuracy~w.r.t training FLOPs by increasing representational
capacity through sparsity. The expanded weight space enables effective
exploration and exploitation by DST algorithms, facilitating the discovery of
optimal sparse subnetworks. Our spectral analysis of Sparse-IFT models trained
with DST reveals efficient connectivity and information propagation, correlated
with high-performance networks. Notably, Sparse-IFT consistently outperforms
dense models in vision and NLP domains. Despite current hardware limitations,
promising benchmarks on Cerebras CS-2 and Neural Magic DeepSparse runtime
highlight the need for improved support for unstructured weight sparsity. We
hope our findings encourage the community to explore unstructured sparsity for
improved model efficiency and performance across applications.
\section*{Acknowledgements}
We thank Anshul Samar for his helpful comments and edits that improved our
manuscript. We also thank Chen-Yu Kevin Leong for assisting with the Cerebras
CS-2 GPT-3 experiments and Dylan Finch for performance evaluations on CS-2.
Finally, we provide details on each author's contributions in
Appendix~\ref{app:authorcontrib}.
\section*{Impact Statement}

The landscape of machine learning (ML) has witnessed an exponential growth in
models, particularly in domains such as natural language processing and computer
vision. However, this surge in model size has come with a considerable cost in
terms of compute, memory, and energy requirements. Our approach, Sparse Iso-FLOP
Transformations (Sparse-IFT), represents a significant stride toward mitigating
these resource-intensive demands. Sparse-IFT introduces a novel approach that
enhances the efficiency of training large neural networks. Remarkably, it
achieves improved accuracy while maintaining the same FLOPs as the original
dense baseline model. Our method holds promise for positive environmental
impacts, given the substantial computational resources typically associated with
training large neural networks.

Models trained with Sparse-IFT require less computing resources and energy to
achieve higher model quality, directly translating to lower deployment costs for
real-world applications. Furthermore, training models with sparsity has been
shown to lead to better generalization~\cite{chen2022sparsity}, a benefit
supported by our transfer learning results on computer vision tasks. An
additional advantage is the enhanced efficiency in training larger sparse
models, facilitated by the widespread adoption of AI hardware, such as the
Cerebras CS-2, which accelerates unstructured sparsity. The key here is
achieving sparsity acceleration through a harmonious collaboration between
hardware support and the development of sparse ML techniques.

The potential sustainability contribution lies in the fact that, as sparse ML
software and hardware co-design continues to evolve, we may be able to train
more accurate ``larger sparse'' networks within the confines of the same
computational budget as a smaller dense model. This paradigm shift could usher
in a more environmentally conscious approach to deep learning, addressing the
concerns associated with the escalating resource requirements of ever-expanding
models. The seamless integration of these elements ensures that hardware
architectures are optimized to complement sparse techniques, fostering a
sustainable and efficient trajectory for the future of deep learning.

As we continue to explore the intersection of hardware support for sparsity and
the evolution of sparse ML techniques, our benchmarking analysis in
Section~\ref{sec:wall_clock} and Appendix~\ref{app:benchmark} serves as a
practical illustration of the transformative potential of Sparse-IFT. It not
only substantiates the theoretical promises but also offers a roadmap for future
developments in the pursuit of sustainable and efficient deep learning
practices.

\bibliography{refs}
\bibliographystyle{icml2024}

\newpage
\appendix
\onecolumn
\section{Additional Methodology Details}

\subsection{Sparse-IFT for Convolutional Layers \label{app:sift_conv}} In this
section, we detail the straightforward extension of the Sparse-IFT family for
convolutional layers.

\paragraph{Sparse Wide}
Similar to the setup for fully connected layers, in the case of convolutional
layers, we widen the number of input and output channels.

\paragraph{Sparse Parallel}
Similar to the setup for fully connected layers, in the case of convolutional
layers, we can implement this transformation with the use of convolutional
branches in parallel.

\paragraph{Sparse Factorized and Sparse Doped}
Let $\theta_l \in \mathbb{R}^{c_{in}\times c_{out} \times k_h \times k_w}$
represent the weight matrix of a convolutional layer, where $c_{in}, c_{out},
k_h, k_w$ denote the input channels, output channels, kernel height, and kernel
width, respectively. We apply low-rank or matrix factorization to the weight
matrix by first converting the 4D tensor into a 2D matrix with shape:
$(c_{in}\cdot k_h \cdot k_w)\times c_{out}$. In this setup, we can express
$\theta_l = UV^T$, where $U \in \mathbb{R}^{c_{in}\cdot k_h \cdot k_w \times
d}$, $V \in \mathbb{R}^{c_{out} \times d}$. In this factorization, $U$ learns a
lower-dimensional set of features and is implemented as a convolutional layer
with $d$ output channels and $k_h \times k_w$ filter. $V$ matrix expands this
low-dimensional set of features and is implemented as a convolutional layer with
$1\times1$ filter.

\subsubsection{Sparse-IFT for Depthwise Convolution Layers}

For a normal convolution layer, all inputs are convolved to all outputs.
However, for depthwise convolutions, each input channel is convolved with its
own set of filters. Let $\theta_l \in \mathbb{R}^{c_{in}\times c_{out} \times
k_h \times k_w}$ represent the weight matrix of a normal convolution layer,
where $c_{in}, c_{out}, k_h, k_w$ denote the input channels, output channels,
kernel height, and kernel width, respectively. An equivalent depthwise
convolution layer will have weights $\theta_{dw,l} \in \mathbb{R}^{ 1\times
c_{out} \times k_h \times k_w}$.

\paragraph{Sparse Wide} A Sparse Wide depthwise convolution will have weights
$\theta_{dw,l}^{sw} \in \mathbb{R}^{ 1\times k_{sw}{\cdot}c_{out} \times k_h
\times k_w}$. Since the fraction of non-sparse weights is given by $1-s$, the
FLOPs required by this transformation are
$B{\cdot}(k_{sw}{\cdot}c_{out}){\cdot}k_h{\cdot}k_w{\cdot}(1-s)$. Setting these
equal to the FLOPs of the original dense $\theta_{dw,l}$, we obtain the widening
factor $k_{sw} = \frac{1}{(1-s)}$. In this case, we do not scale the input
channels as it converts the depthwise convolution to a grouped convolution
without an equivalent scaling in the number of groups.

\paragraph{Other Sparse-IFT Transformations} The Sparse Wide IFT generally
changes a layer's input and output channels, subsequently scaling the following
layers in a CNN. However, the other Sparse-IFT transforms (Sparse Parallel,
Sparse Factorized, and Sparse Doped) do not modify a convolution layer's input
or output channels (as seen in Figure~\ref{fig:diff_members_of_sift}). This
allows for fine-grained control of what layers to apply the Sparse-IFT
transformations. Since depthwise convolutions are an extreme form of structured
sparsity, where some filters interact with only specific input channels, we opt
not to sparsify them when using the other Sparse-IFT transformations and leave
the layer unchanged while still maintaining FLOPs equivalent to the dense
baseline. Note that the different convolution layers surrounding the depthwise
convolution are still transformed with Sparse-IFT to increase their
representational capacity.

\subsection{Controlling for Iso-FLOP}
\label{app:control_flops}
As mentioned before, in our work, we mainly apply a uniform sparsity
distribution to the model, which essentially means each layer is allocated the
same level of sparsity.  Let $\mathcal{N}$ denote a $L$ layered DNN
parameterized by $\Theta_{\mathcal{N}}$. Let $\Theta_{\mathcal{N}} \in
\{\theta_1, \ldots, \theta_L\}$ denote the parameters of the DNN. Now, let $M_l$
be the binary mask for layer $l \in \{1,\ldots,L\}$ with dimensions
corresponding to the parameters of that layer. The binary mask $m_l$ has values
of 1 for active weights and 0 for non-active weights. Let $\theta_l$ be the
total number of parameters in $l$, hence, the sparsity level per layer, $s_l$,
is defined as $\frac{\sum_{i,j} \mathbb{I} (m_l(i,j) \neq 0)}{\theta_l}$. The
average sparsity level in the network, $s$, is then defined as the ratio of the
total number of zero parameters to the total number of parameters. This is
expressed as $s = \frac{\sum_{l=1}^{L}\sum_{i,j}\mathbb{I}(m_l(i,j) \neq
0)}{\Theta_{\mathcal{N}}}$. Below, we characterize the different scenarios when
training with different sparse training methods:

 \begin{itemize}
    \item \textbf{Random Static Sparsity:} In this case, the sparsity
    distribution is uniform, ensuring that the sparsity in each layer matches
    the target sparsity level. Consequently, the application of Sparse-IFT,
    parameterized by the sparsity level, maintains Iso-FLOP equivalence to the
    original dense model. However, adhering to common practice for computer
    vision networks (e.g., ResNet), we retain the first and last layers (input
    convolution and output linear layer) as dense to prevent a significant
    decline in model quality during pre-training. Consequently, the Sparse-IFT
    network deviates from Iso-FLOP to the dense model, introducing additional
    FLOPs that need consideration.
    \item \textbf{Pruning at Initialization:} The algorithms, such as
    SNIP~\citep{lee2018snip}, GraSP~\citep{wang2020picking},
    FORCE~\citep{de2020progressive}, etc., introduce distinct criteria or
    methods for determining which weights to prune at initialization,
    influencing the sparsity distribution. Consequently, the inherent
    characteristics of these algorithms result in changes to the sparsity
    distribution. In the context of Sparse-IFT, despite having an identical
    total sparse parameter count to the original dense model, the Sparse-IFT
    network no longer maintains Iso-FLOP equivalence.
    \item \textbf{Dense-to-Sparse Training:} Sparse training methods, such as
    GraNet~\citep{liu2021sparse}, employ dense-to-sparse training, initiating
    training from either a fully dense state or a state less sparse than the
    target sparsity level. For instance, GraNet utilizes gradual magnitude
    pruning~\citep{Zhu2017ToPO} at the beginning of training to systematically
    reduce the network's density to the target sparsity level. Consequently, in
    the context of Sparse-IFT networks, this configuration no longer maintains
    Iso-FLOP equivalence to the dense model, as the average training FLOPs
    surpass those of the original dense model.
 \end{itemize}

 To address the FLOPs discrepancy between the Sparse-IFT network trained with
 non-uniform sparsity distributions (e.g., PaI methods or densifying certain
 layers) and dense-to-sparse training (e.g., GraNet), we employ a binary search
 to fine-tune the target sparsity of the network prior to any training. In this
 process, we set the maximum and minimum values for the target sparsity level.
 At each iteration, we profile the FLOPs used by the Sparse-IFT network and
 compare it to the original dense model FLOPs. The target sparsity level is
 adjusted through the binary search, ensuring that the total FLOPs of the
 Sparse-IFT network are within 0.0001\% of the dense model FLOPs.

\section{Graph Analysis of Sparse-IFT with DST}
\label{app:graph_analysis_sift_dst}
In our analysis, we interpret the Sparse-IFT ResNet-18 models as a series of
bipartite compute graphs, where each layer, $\{\theta_1,\ldots, \theta_L\}$ in
an $L$ layered sparse DNN,  takes the form of a square adjacency matrix $A$. The
Ramanujan gap is defined as $\Delta r = 2 \ast \sqrt{d-1} -
\hat{\mu}(A)$~\citep{hoang2023revisiting,hoang2023dont}, where $d$ is the
average edge per node, and $\hat{\mu}(A)$ is the non-trivial eigenvalue of $A$.
Also, we analyze the Iterative Mean Difference Bound, $\Delta r_{imdb} =
\frac{1}{\abs{K}} \sum_{i=1}^{\abs{K}}(2\sqrt{d_i - 1} -
\hat{\mu}(A_{K_i}))$~\citep{hoang2023revisiting}. We train a ResNet-18 model
with all members of the Sparse-IFT family using a dynamic sparse training
algorithm (i.e., RigL~\citep{evci2020rigging}). 

\begin{figure*}
    \vspace{-0.1in}
    \centering
    \subfigure{\includegraphics[width=0.30\textwidth]{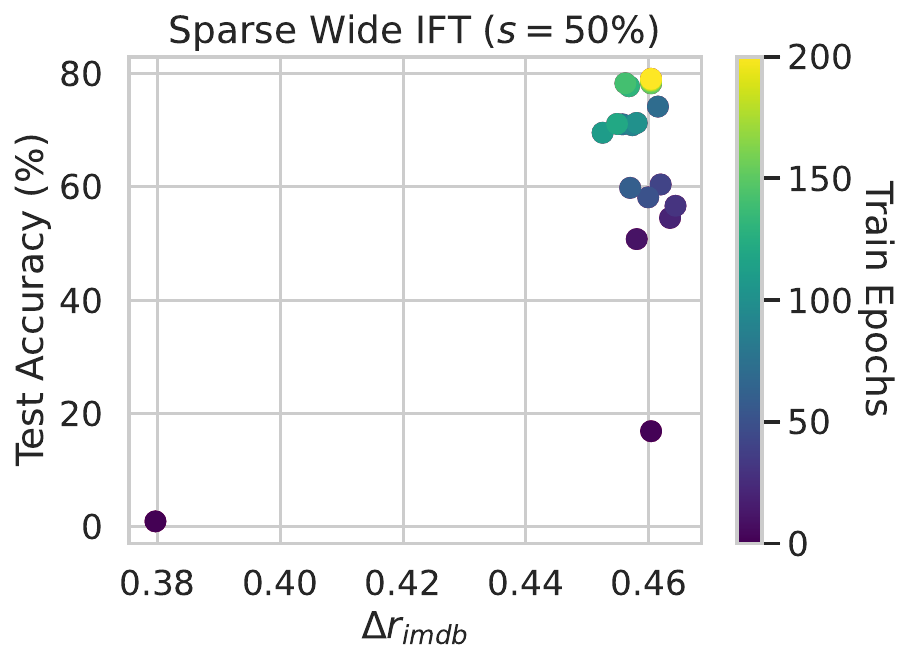}}
    \subfigure{\includegraphics[width=0.30\textwidth]{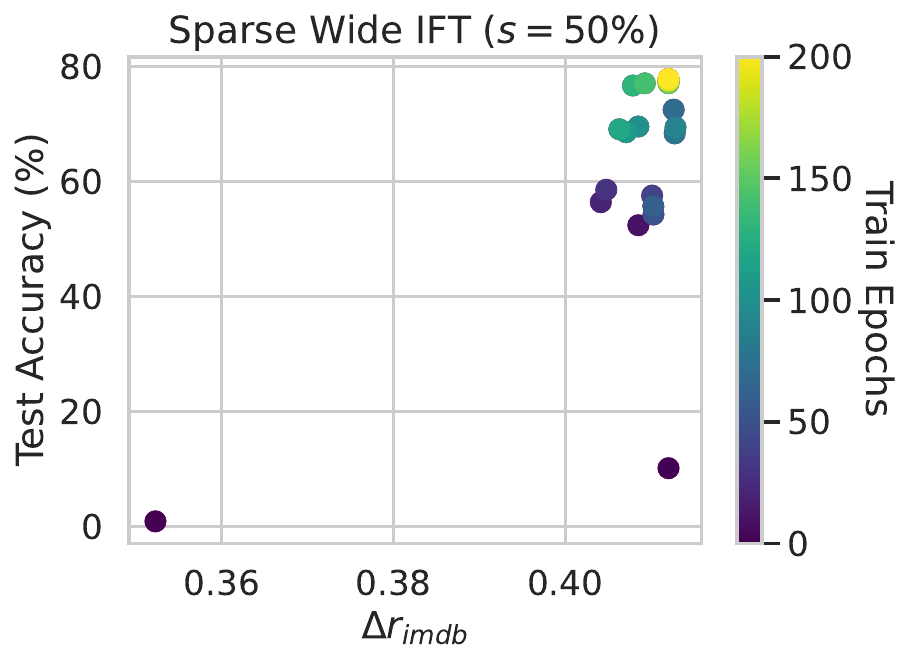}}
    \subfigure{\includegraphics[width=0.30\textwidth]{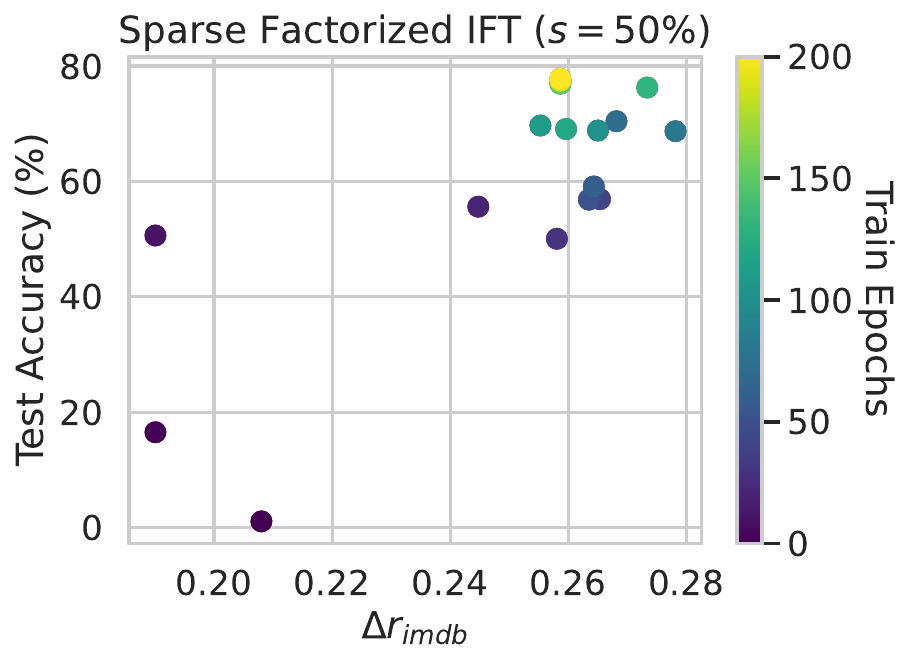}}
    \\ 
    \subfigure{\includegraphics[width=0.30\textwidth]{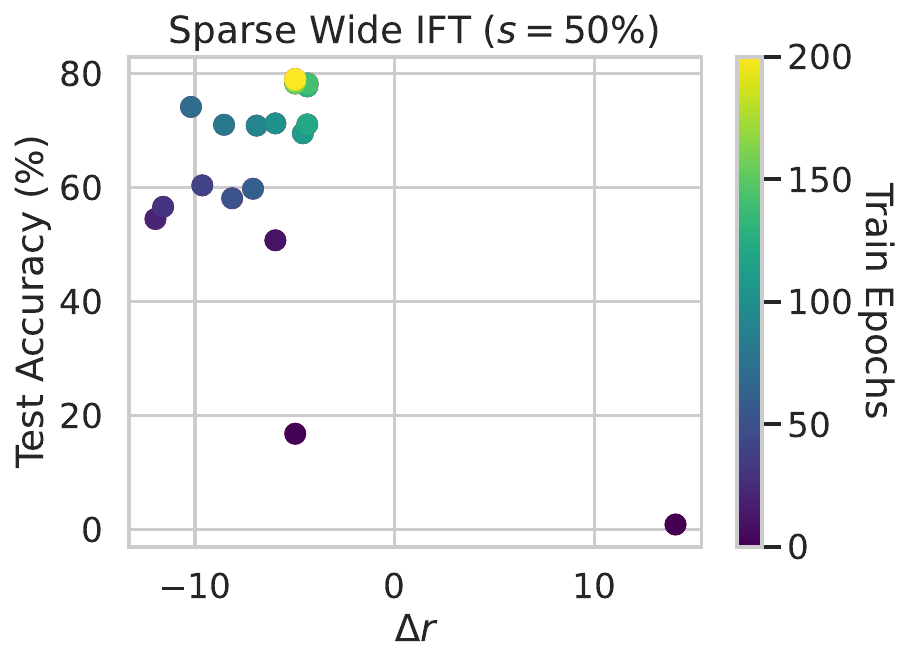}}
    \subfigure{\includegraphics[width=0.30\textwidth]{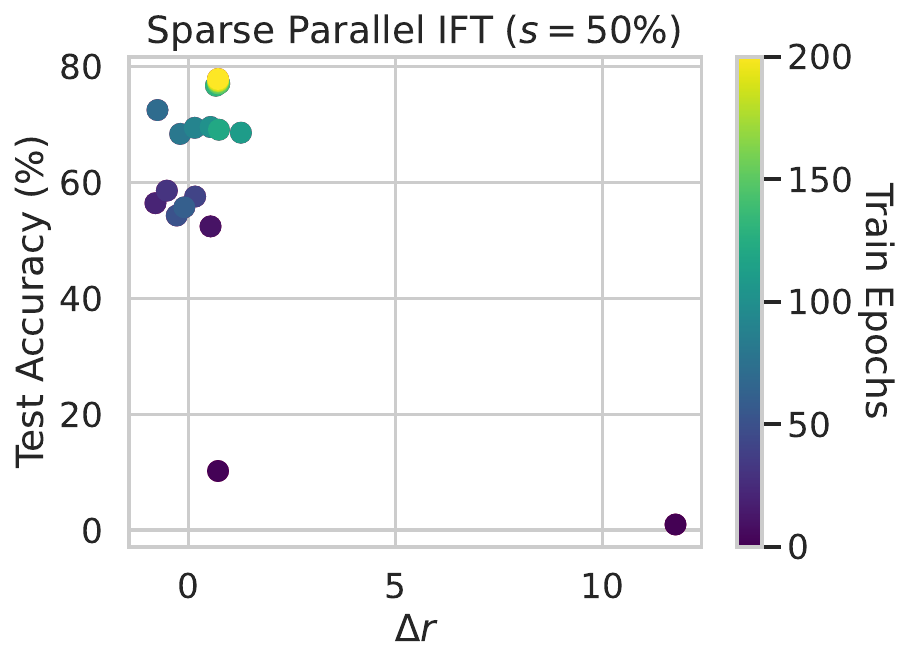}}
    \subfigure{\includegraphics[width=0.30\textwidth]{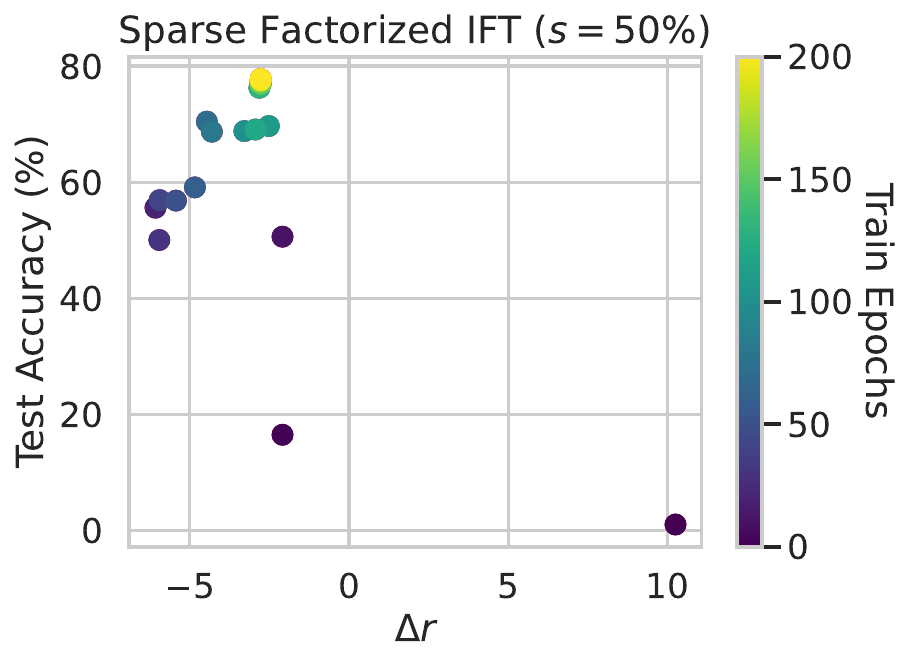}}
    \vspace{-15pt}
    \caption{Top row illustrates the dynamic interplay between the Iterative
    Mean Difference Bound, $\Delta r_{imdb}$ and test accuracy, and bottom row
    shows the correlation between the Ramanujan Gap, $\Delta r$ and test
    accuracy throughout the training process. This illustrates the evolving
    relationship between spectral graph properties and network performance,
    shedding light on the connectivity dynamics of the Sparse-IFT networks
    trained with DST.}
    \vspace{-5pt}
    \label{fig:graph_analysis_detailed}
    \vspace{-10pt}
\end{figure*}

\textbf{$\vb*{\Delta r}$ Analysis:} In Figure~\ref{fig:graph_analysis_detailed},
we observe that $\Delta r$ decreases over the course of training and then
maximizes at later stages, which suggests that the spectral properties of the
adjacency matrices are changing dynamically during training. The fact that
$\Delta r$ maximizes at later stages and correlates with the Sparse-IFT
ResNet-18 model achieving the highest test accuracy indicates a potential
connection between the spectral properties of the adjacency matrices and the
model's performance. The dynamic changes in $\Delta r$ might indicate that the
neural network is adapting its structure during training. The network might be
pruning less important connections and reinforcing more important ones, leading
to an optimized structure. Moreover, the increase in $\Delta r$ could be related
to implicit regularization effects. The spectral properties of the adjacency
matrices may play a role in controlling the model's capacity, preventing
overfitting, and enhancing generalization. The correlation between the
maximization of $\Delta r$ at the later stages of training and the highest test
accuracy suggests that there is a relationship between the identified spectral
properties and the performance of the Sparse-IFT ResNet-18 model. The
maximization of $\Delta r$ could represent an optimal point in the trade-off
between sparsity and model accuracy for the given task.

\textbf{$\vb*{\Delta r_{imdb}}$ Analysis:} The increasing trend of $\Delta
r_{imdb}$ during training suggests that the overall connectivity boundary across
subgraphs is progressively being enhanced. This could imply that the network is
learning to establish more meaningful and relevant connections within its
structure as training progresses. The DST algorithm may be facilitating an
adaptive refinement of connectivity within the network. The observed increase in
$\Delta r_{imdb}$ could indicate that the model is iteratively adjusting its
connectivity boundaries to improve information flow. $\Delta r_{imdb}$ evaluates
the average connectivity boundary across all subgraphs, providing a more
comprehensive measure of the network's overall connectivity changes. The
correlation with the highest performing models at the final stage of training
suggests that the average connectivity enhancements captured by $\Delta
r_{imdb}$ are beneficial for the model's performance.

\section{Computer Vision: Experimental Settings \label{app:cv-exp-details}}

\subsection{Computer Vision: Pre-Training Settings \label{app:pt-implement}}
\paragraph{CIFAR-100} Our implementation of CIFAR-100 follows the setup
from~\cite{devries2017improved} for ResNets. We train the models for 200 epochs
with batches of 128 using SGD, Nesterov momentum of 0.9, and weight-decay of
5$\times 10^{-4}$. The learning rate is initially set to 0.1 and is scheduled to
decay to decrease by a factor of 5x after each of the 60th, 120th, and 160th
epochs. Following recent advances in improving ResNets, we initialize the
network with Kaiming He initialization~\cite{he2016identity}, zero-init
residuals~\cite{he2019bag}, and disable weight-decay in biases and
BatchNorm~\cite{ioffe2015batch} layers. For CIFAR-100 experiments with
MobileNetV2, MobileViT-S, and BotNet-50, we follow the same training setup used
for ResNet, but the learning rate is scheduled via cosine annealing.

\paragraph{ImageNet} Our implementation of ImageNet follows the standard setup
from~\cite{krizhevsky2012imagenet, simonyan2014very}. The image is resized with
its shorter side randomly sampled in [256, 480] for scale
augmentation~\cite{simonyan2014very}. A 224 $\times$ 224 crop is randomly
sampled from an image or its horizontal flop, and then normalized. For
evaluation, the image is first resized to 256 $\times$ 256, followed by a 224
$\times$ 224 center crop, and then normalized. Following recent advances in
improving ResNets, we initialize the network with Kaiming He
initialization~\cite{he2016identity} and zero-init residuals~\cite{he2019bag}.

For ResNets, we replicate the settings recommended by Nvidia~\cite{resnetv15},
which uses the SGD optimizer with a momentum of 0.875 and weight decay of
3.0517578125$\times 10^{-5}$. We disable weight-decay for biases and BatchNorm
layers. The model is trained with label smoothing~\cite{szegedy2016rethinking}
of 0.1 and mixed precision~\cite{micikevicius2017mixed} for the standard 90
epochs using a cosine-decay learning rate schedule with an initial learning rate
of 0.256 for a batch size of 256.~\citet{srinivas2021bottleneck} follow the same
setup as ResNet for training BotNet-50 on ImageNet, therefore we maintain the
same hyperparameter settings as~\citet{resnetv15} for our BotNet-50 ImageNet
experiments.

\paragraph{Sparsity Setup} For enabling the Sparse-IFT transformations, we use
the RigL~\cite{evci2020rigging} algorithm in its default hyperparameter settings
($\alpha = 0.3, \Delta T = 100$), with the drop-fraction ($\alpha$) annealed
using a cosine decay schedule for 75\% of the training run. We keep the first
and last layers (input convolution and output linear layer) dense to prevent a
significant degradation in model quality during pre-training, which is standard
practice. We account for these additional dense FLOPs by increasing the sparsity
in the remaining layers, similar to~\citet{gale2019state}
and~\citet{liu2022unreasonable}.

\subsection{Importance of Non-linearity}\label{app:nonlinear-importance} We use
BatchNorm~\cite{ioffe2015batch} followed by ReLU~\cite{nair2010rectified} as a
non-linearity. We provide an extended set of empirical results in
Table~\ref{tab:non_linearity_importance} to help validate the importance of
training with and without non-linearity by training configurations of the Sparse
Parallel, Factorized, and Doped IFT families at different levels of sparsity.
The results without non-linear activation functions are often worse than the
dense accuracy (77\%) across all Sparse-IFT family transformations. We omit
Sparse Wide in Table~\ref{tab:non_linearity_importance} because here we increase
the number of channels in the convolutional layers while maintaining the
existing architecture.

\begin{table}[h]
    \caption{Evaluation on the importance of utilizing the non-linear activation
across different members of Sparse-IFT with ResNet-18 on CIFAR100 across
different values of sparsity (columns). Non-linear activations enhance the
representational capacity of Sparse-IFT, leading to higher accuracy. All
reported results are the average over 3 random seeds.} 
    \centering
    \begin{small}
    \begin{tabular}{c|cccc}
        \toprule
        Transformation & Non-linear activation & 0.50  & 0.75 & 0.90  \\
	\midrule
    \multirow{2}{*}{Sparse Factorized} & \xmark    &  75.9 $\pm$ 0.3 & 76.6
    $\pm$ 0.4 & 76.5 $\pm$ 0.4  \\
       &  \cmark

&  \textbf{77.8 $\pm$ 0.4} &  \textbf{78.4 $\pm$ 0.5} & \textbf{78.9  $\pm$ 0.5}
\\
       \midrule \multirow{2}{*}{Sparse Parallel} & \xmark            	      &
       77.1 $\pm$ 0.1 &  77.2 $\pm$ 0.2 & 77.6 $\pm$ 0.1  \\
       &  \cmark         	      &  \textbf{77.9 $\pm$ 0.2 } &  \textbf{79.1
       $\pm$ 0.2 } & \textbf{78.2 $\pm$ 0.2} \\
       \midrule \multirow{2}{*}{Sparse Doped} & \xmark            	      & 77.3
       $\pm$ 0.2 &  77.1 $\pm$ 0.1 & 76.5 $\pm$ 0.2  \\
       &  \cmark         	      &  \textbf{78.2 $\pm$ 0.1} &  \textbf{77.8
       $\pm$ 0.1} & \textbf{76.9 $\pm$ 0.2} \\
        \bottomrule
    \end{tabular}
    \end{small}
    \label{tab:non_linearity_importance}
\end{table}

\subsection{Computer Vision \label{app:ft-implement}}

\subsubsection{Sparse-IFT vs. Extended Sparse Training Schedules}
\label{app:sift_vs_extendedsparse}
We provide a direct comparison with sparse training methods (e.g., RigL and SET)
in the Iso-FLOP setting (i.e., training with a longer schedule) to demonstrate
the significance of our results with respect to this standard sparse baselines.
As shown in the Table~\ref{tab:sift_vs_extendedsparse}, Sparse-IFTs outperform
dynamic sparse training methods by a significant margin across all levels of
sparsity. Note, at higher levels of sparsity (e.g., 90\%), sparse training
methods obtain worse accuracy compared to the FLOP equivalent dense baseline. In
contrast, with Sparse-IFT, we observe higher accuracy across all levels of
sparsity evaluated.

\begin{table}
    \caption{Results with ResNet-18 on CIFAR-100 across different values of
    sparsity (columns). Best accuracy for each sparse training method is
    highlighted in bold. The original dense ResNet-18 model obtains an accuracy
    of 77.0$\pm$0.2. All reported results are over 3 random seeds. } 
    \centering
    \begin{small}
    \resizebox{\columnwidth}{!}{
    \begin{tabular}{c|cccccc}
        \toprule
        Dense 		& Transformation & Sparse Training Method & Epochs & 0.50  &
        0.75 & 0.90  \\
	\midrule
 \multirow{4}{*}{77.0 $\pm$ 0.2}
		& Sparse Wide &  SET & 200 $\cdot$ $\frac{1}{1-s}$ & \textbf{78.7 $\pm$
		0.2} & 78.4 $\pm$ 0.1 & 76.8 $\pm$ 0.1  \\
		&  Sparse Wide  &  RigL & 200 $\cdot$ $\frac{1}{1-s}$ & \textbf{78.9
		$\pm$ 0.1} & 78.8 $\pm$ 0.1 & 76.4 $\pm$ 0.2 \\
		&  Sparse Wide  &  RigL  & 200 &  79.1 $\pm$ 0.2 & 79.5 $\pm$ 0.1 &
		\textbf{80.1 $\pm$ 0.2} \\
        \bottomrule
    \end{tabular}
    }
\end{small}
\label{tab:sift_vs_extendedsparse}
\end{table}

\subsubsection{Sparse-IFT on Efficient Computer Vision
Architectures}\label{app:efficientcv} Here, we provide an extended set of
results on MobileNetV2, MobileViT-S, and BotNet-50 on CIFAR-100. In particular,
we enable Sparse Wide and Sparse Parallel IFT at 50\% and 75\% sparsity values
(see Table~\ref{tab:mbv2-cifar-app}).

\begin{table}[!ht]
    \caption{Evaluation of Sparse Wide and Sparse Parallel IFT with various
compute efficient architectures on CIFAR-100 across different values of sparsity
(columns). Using Sparse Parallel IFT, all architectures outperform the dense
baseline by a significant margin.} \vskip 2pt
    \centering
    \begin{small}
    \begin{tabular}{c|c|ccc}
        \toprule
        	    & Dense  & Transformation           & 0.50 & 0.75 \\ \midrule
                \multirow{2}{*}{MobileNetV2} & \multirow{2}{*}{72.4 $\pm$ 0.2 }
                & Sparse Wide  & 73.4 & \textbf{73.7} \\

                & & Sparse Parallel & 72.9 & \textbf{73.3} \\ \midrule
                \multirow{2}{*}{MobileViT-S} & \multirow{2}{*}{73.5 $\pm$ 0.1} &
                Sparse Wide & 74.6 & \textbf{74.8} \\
                & & Sparse Parallel & 73.7 & \textbf{74.4} \\ \midrule
\multirow{2}{*}{BotNet-50}  & \multirow{2}{*}{79.8 $\pm$ 0.2} & Sparse Wide &
80.3 & \textbf{80.6} \\
        & & Sparse Parallel & 79.7 & \textbf{80.5} \\
        \bottomrule
    \end{tabular}
    \end{small}
\label{tab:mbv2-cifar-app}
\end{table}

\subsubsection{Evaluation of Sparse-IFT with Structured Sparsity}
\label{app:structured_sparsity}

\paragraph{Block Sparsity}
To derive Iso-FLOP configurations with block sparsity, we reuse the analysis
done previously with unstructured sparsity (see
Section~\ref{subsec:members_of_sift}) and express the width scaling as a
function of sparsity. However, we will search for a block sparse mask during
training instead of an unstructured sparsity mask. We use the method proposed
by~\citet{hubara2021accel} to search N:M transposable sparsity, which can
accelerate both the forward and backward pass during training on NVIDIA GPUs
with Tensor Cores. We use 4:8-T, 2:8-T, and 1:8-T block patterns to obtain 50\%,
75\%, and 87.5\% sparsity, respectively. Note the 1:8-T block is the closest
approximation to a 90\% sparsity pattern attainable with a block size of 8. We
also set up and experimented using the method proposed
by~\citet{jiang2022exposing} to train with fine-grained sparse block structures
dynamically. However, the algorithm uses agglomerative clustering which led to a
much slower runtime and quickly ran out of memory even at 50\% sparsity using
the Sparse Wide IFT on a single Nvidia V100 (16 GB).

\paragraph{Low Rank}
Let $k_{lr}$ be the factor with which we widen all layers' input and output
 dimensions for low-rank factorization. We replace all dense layers with
 low-rank factorization, i.e. $\theta_l^{lr} = U_lV_{l}^{T}$, where $U_l \in
 \mathbb{R}^{(k_{lr}.D_{in}) \times d}$ and $V_l \in \mathbb{R}^{
 (k_{lr}.D_{out}) \times d}$. Given a widening factor and equating the FLOPs of
 this transformation to that of a dense transformation $f_{\theta}$, we obtain
 the following expression for rank $d$: $\frac{D_{in}.D_{out}.k_{lr}}{(D_{in} +
 D_{out}}$. We evaluate this factorization across different values of
 width-scaling $k_{lr}$ in Table~\ref{tab:struct-cifar-lowrank}.

\begin{table*}[]
    \caption{Comparison of structured sparse and unstructured sparse methods on
    CIFAR-100 test accuracy on ResNet-18.} 
    \centering
    \begin{small}
    \begin{tabular}{c|cccccc}
        \toprule
        &               &          & \multicolumn{4}{l}{Width Scaling Factor} \\
        Transformation & Sparsity Type & Sparsity & 1x      & 1.41x     & 2x &
        3.16x     \\ \midrule Low Rank, Linear &  Structured & 0\% & 74.1 & 74.3
        & 74.3 & 73.4 \\
        Low Rank, Non-Linear & Structured  &  0\% &  76.8 & 76.5 & 76.0 & 75.3
        \\
\midrule
        \multirow{6}{*}{Sparse Wide} &
\multirow{3}{*}{\begin{tabular}[c]{@{}c@{}}N:M Block Sparse\\
\citep{hubara2021accel}\end{tabular}} & 4:8-T    &         &    77.1       & &
 \\
        &                                       & 2:8-T    &         & &
        \textbf{78.4} &           \\
        &                                       & 1:8-T    &         & & & 78.1
        \\
   & \multirow{3}{*}{\begin{tabular}[c]{@{}c@{}}Unstructured Sparse\\
\citep{evci2020rigging}\end{tabular} } & 50\%    &         &    79.1       & &
\\
        &                                       & 75\%   &         &           &
        79.5     &           \\
        &                                       & 90\%    &         & &        &
        \textbf{80.1}  \\
        \bottomrule
    \end{tabular}
    \end{small}
\label{tab:struct-cifar-lowrank}
\end{table*}

\subsubsection{Evaluation on downstream tasks \label{app:eval_downstream}}
\subsubsection*{COCO Object Detection \label{app:coco}} This dataset contains
118K training, 5K validation (\texttt{minival}), and 20K test-dev images. We
adopt the standard single-scale training setting~\cite{lin2017feature} where
there is no additional data augmentation beyond standard horizontal flipping.
For training and testing, the input images are resized so that the shorter edge
is 800 pixels~\cite{lin2017feature}. The model is trained with a batch size of
16, using the SGD optimizer with a momentum of 0.9 and weight decay of 1$\times
10^{-4}$. We follow the standard 1x schedule (12 epochs) using a step learning
rate schedule, with a 10x decrease at epochs 8 and 11, an initial learning rate
warmup of 500 steps starting from a learning rate of 2$\times 10^{-5}$, and a
peak learning rate of 0.01.

\begin{table*}[h]
    \caption{ Object detection results on COCO \texttt{minival} in the RetinaNet
        framework. Sparse Wide IFT configurations of RetinaNet outperform the
        dense baseline by a large margin on all metrics while using similar
        FLOPs. } 
    \centering
    \begin{small}
    \begin{tabular}{c|cccccccc}
        \toprule
        Backbone   &      AP   & AP$_{50}$ & AP$_{75}$ & AP$_{S}$ & AP$_{M}$ &
        AP$_{L}$ \\ \midrule Dense           &   29.3 & 46.2 & 30.9 & 14.7 &
        31.5 & 39.6 \\
        Sparse Wide (50\%)  & 31.3 & 49.0 & 33.0 & 16.6 & 34.0 & 42.0 \\
        Sparse Wide (75\%) & 32.8 & 51.0 & 34.8 & 17.3 & 35.8 & 43.3 \\
        Sparse Wide (90\%) & \textbf{34.5} & \textbf{53.5} & \textbf{36.5} &
        \textbf{18.6} & \textbf{37.6} & \textbf{45.3} \\
        \bottomrule
    \end{tabular}
    \end{small}
    \label{tab:app-cocoretinanet}
\end{table*}

\subsubsection*{CityScapes Semantic Segmenation \label{app:cityscapes}}

\paragraph{Setup} We follow the same training protocol
as~\cite{zhao2017pyramid}, where the data is augmented by random cropping (from
1024 $\times$ 2048 to 512 $\times$ 1024), random scaling in the range [0.5, 2],
and random horizontal flipping. The model is trained with a batch size of 16,
using the SGD optimizer with a momentum of 0.9 and weight decay of 5$\times
10^{-4}$. We follow the 80K iterations setup from MMSegmentation with an initial
learning rate of 0.01 annealed using a poly learning rate schedule to a minimum
of 1$\times 10^{-4}$. Similar to most setups that tune
hyperparameters~\cite{zhao2017pyramid, liu2021swin, wang2020deep} for reporting
the best results, we tune the learning rate for all our models. All our results
are reported using a learning rate of 0.03 for the sparse backbones and 0.01 for
the dense baseline.

\begin{table*}[h]
    \caption{ Semantic segmentation results on the Cityscapes \texttt{val} set
        using DeepLabV3+. Sparse Wide IFT configurations ResNet-18 backbones
        outperform the dense baseline on all metrics while using similar FLOPs.
        }
    \centering
    \begin{small}
    \begin{tabular}{c|cc}
        \toprule
        Backbone  &  mIoU & mAcc \\ \midrule Dense          & 76.72 & 84.40 \\
        Sparse Wide (50\%)      &  77.90 & 85.12 \\
        Sparse Wide  (75\%)      &   78.92 & 85.68 \\
        Sparse Wide  (90\%)      &  \textbf{79.10} & \textbf{86.01} \\
        \bottomrule
    \end{tabular}
    \end{small}
    \label{tab:app-cityscapes}
\end{table*}

\section{Natural Language Processing: Experimental
Settings~\label{app:llm-setup}}

\subsection{Details for GPT End-to-End Training}
\label{app:gpt_e2e}
We demonstrate the benefits of using Sparse-IFT transformations in the NLP
domain by pre-training GPT-3 models and performing zero-shot eval on downstream
tasks from the HuggingFace Open LLM leaderboard. Here, we pre-train the models
on the Pile~\cite{gao2020pile} dataset. To train all GPT models, we use the
AdamW optimizer \cite{loshchilov2017decoupled} with $\beta_1 = 0.9$, $\beta_2 =
0.95$ and $\epsilon = 10^{-8}$. The global norm is clipped at 1.0, and a weight
decay of 0.1 is used. There is a learning rate warmup over the first 375M
tokens, followed by a cosine decay to 10\% of the peak learning rate. We follow
the recently published Chinchilla \cite{hoffmann2022an} recommendations for
obtaining loss-optimal pre-trained baseline configurations of models. The
context window size is 2048 following \cite{brown2020language}.
Table~\ref{tab:app-gpt-pt} shows a detailed breakdown of the model
architectures, learning rate, and training settings. 

In Tables~\ref{tab:app-gpt-pt} and~\ref{tab:app-gpt-pt-sift}, we outline the
architecture configurations for Sparse Wide IFT 50\% and 75\% variants. We train
the Sparse Wide GPT-3 models using the dynamic sparse training algorithm,
SET~\citep{mocanu2018} on the Cerebras CS-2 to realize the acceleration from
unstructured sparsity. Currently, Cerebras CS-2’s specialized kernels support
training with dynamic unstructured sparsity via SET; therefore, results in this
section are reported with SET. In Table~\ref{tab:app-gpt-pt}, $n_{params}$ is
the total number of trainable parameters, $n_{layers}$ is the number of decoder
layers, and $d_{model}$ is the base size of the model. The feedforward
bottleneck is four times the base size, i.e., $d_{\text{ff}} = 4\times
d_{model}$. Finally, $n_{heads}$ is the number of attention heads, and
$d_{head}$ is the dimension of each attention head. 
\begin{table*}[]
    \caption{ Size, architecture, and learning hyperparameters (batch size and
        learning rate) of the GPT-3 Small model, which is trained using
        Chinchilla optimal configurations ($\approx$ 20 tokens per parameter) }
    \centering
    \resizebox{\textwidth}{!}{
    \begin{small}
    \begin{tabular}{c|ccccc|cc|c}
        \toprule
        Model & $n_{params}$ & $n_{layers}$ & $d_{model}$ & $n_{heads}$  &
        $d_{head}$ & Batch Size & Learning Rate & Training Tokens \\ \midrule
        GPT-3 Small           &  125M & 12 & 768 & 12 & 64 & 256 & 6$\times
        10^{-4}$ & 2.5B \\
        \bottomrule
    \end{tabular}
    \end{small}
    }
    \label{tab:app-gpt-pt}
\end{table*}

\begin{table*}[]
    \caption{ Sizes and architecture definitions of the dense GPT-3 Small model
        and its Sparse Wide IFT variants. } 
    \centering
    \begin{small}
    \begin{sc}
    \begin{tabular}{c|c|c|ccccc}
        \toprule
        Model & Transformation & Sparsity & $n_{layers}$ & $d_{model}$ &
        $d_{\text{ff}}$ & $n_{heads}$  & $d_{head}$  \\ \midrule GPT-3 Small &
        Dense   & 0\% & 12 & 768 & 3072 & 12 & 64 \\
        GPT-3 Small & Sparse Wide   & 50\% & 12 & 1092 & 4344 & 12 & 64 \\
        GPT-3 Small & Sparse Wide   & 75\% & 12 & 1536 & 6144 & 12 & 64 \\
        \bottomrule
    \end{tabular}
    \end{sc}
    \end{small}
    \label{tab:app-gpt-pt-sift}
\end{table*}
\paragraph{Evaluation} We conducted a comprehensive evaluation of both dense and
Sparse Wide IFT GPT-3 Small models, assessing their performance at 50\% and 75\%
sparsity levels across five distinct tasks on the Open LLM
leaderboard~\citep{open-llm-leaderboard} using the
LM-eval-harness~\citep{eval-harness}. The tasks encompassed
ARC~\citep{clark2018think}, HellaSwag~\citep{zellers2019hellaswag},
TruthfulQA~\citep{lin2022truthfulqa}, MMLU~\citep{hendrycks2021measuring}, and
Winogrande~\citep{DBLP:journals/corr/abs-1907-10641}. In
Table~\ref{tab:eleuther}, our results reveal that the Sparse IFT GPT-3 Small
model at 75\% sparsity achieved a notable 0.9\% improvement over the dense
baseline, underscoring the efficacy of Sparse Wide IFT in enhancing model
performance across a diverse range of language understanding tasks.

\begin{table}[!h]
    \caption{Performance Evaluation of Dense and Sparse Wide IFT GPT-3 Small
    Models at 50\% and 75\% sparsity levels across five tasks (i.e., ARC,
    HellaSwag, TruthfulQA, MMLU, and Winogrande) on the Open LLM Leaderboard}
    \begin{center}
    \begin{small} 
    \begin{sc}
    \resizebox{\linewidth}{!}{
    \begin{tabular}{c|c|c|c|cccccc}
        \toprule
    \multirow{2}{*}{Model} & \multirow{2}{*}{Transformation} &
    \multirow{2}{*}{Sparsity} & \multirow{2}{*}{Sparse Method} &
    \multicolumn{6}{c}{Open LLM Leaderboard}                    \\
                           &                                 & & & ARC  &
                           HellaSwag & TruthfulQA & MMLU & Winogrande & Average
                           \\ \midrule \multirow{3}{*}{GPT-3 Small} & Dense &
                           0\% &                -                & 20.8 & 27.2 &
                           47.0         & 24.6 & 49.4       & 33.8    \\
                           & Sparse Wide                     & 50\% & SET & 20.6
                           & 27.4      & 47.4      & 25.6 & 49.6 & 34.1    \\
                           & Sparse Wide                     & 75\% & SET &
                           \textbf{22.1} & \textbf{27.8}      & \textbf{47.5} &
                           \textbf{25.6} & \textbf{50.4}       & \textbf{34.7}
                           \\ \bottomrule
    \end{tabular}
    }
    \end{sc}
    \end{small}
    \end{center}
    \label{tab:eleuther}
\end{table}

\section{Benchmarking Efficiency w.r.t Wall-clock \label{app:benchmark}} In this
section we provide additional details on the benchmarking setups for inference
on Neural Magic
DeepSparse~\citep{neural_magic_2021,neuralmagic2021,pmlrv119kurtz20a}
sparsity-aware runtime and training on the Cerebras CS-2~\citep{lie2023cerebras,
lie_2023} for evaluating the efficiency of Sparse-IFT with respect to the
wall-clock time. 

\vspace{-0.1in}

\begin{figure*}[!ht]
    \centering
    \includegraphics[keepaspectratio=true, width=0.45\linewidth]{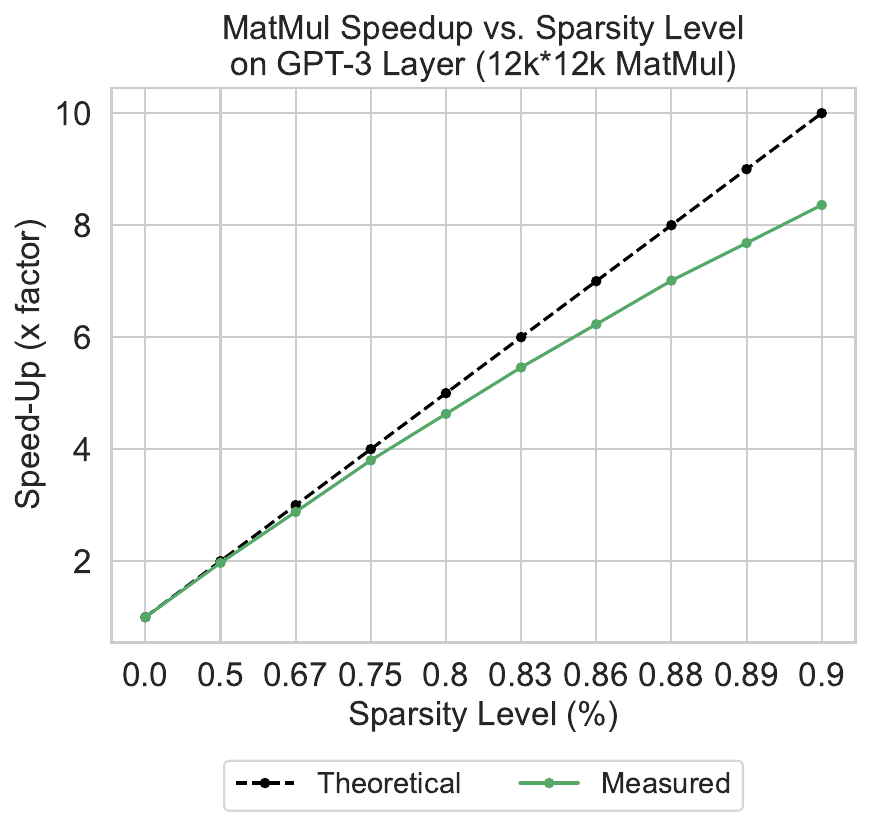}
    \caption{Measured speedup versus theoretical speedup at varying sparsity
    levels for a GPT-3 layer 12k $\times$ 12k matrix multiplication
    (MatMul)~\citep{lie_2021}.}
    \label{fig:matmulspeedup}
\end{figure*}

\paragraph{Inference Setup}
We use Neural Magic's DeepSparse tool for benchmarking Sparse-IFT variants. The
benchmarking is conducted on the Intel Cascade Lake CPUs found on AWS G4dn cloud
instances. These instances support the AVX-512 instruction set, which is used by
the DeepSparse inference runtime to accelerate unstructured sparsity. We
benchmark different configurations of the Sparse Wide ResNet-18 model with
sparsity $\in$ \{50\%, 75\%, 90\%\} for batched inference on ImageNet. We report
runtime for batch-inference of 64 images at 224 $\times$ 224 resolution.

\paragraph{Training Setup}
We evaluate the training efficiency of Sparse-IFT on the Cerebras CS-2 which
supports and accelerates training with unstructured sparsity (both forward and
backward passes). We benchmark the training speed measured in seconds/iteration.
Note that the overall FLOPs of models in the GPT family are comprised of matrix
multiplication FLOPs and attention FLOPs. Attention FLOPs (i.e., spent in
multi-head attention) scale quadratically with sequence length and are invariant
to weight sparsity. To demonstrate the efficacy of sparse kernels for
unstructured weight sparsity, we report our results for dense and Sparse Wide
variants of the GPT-3 20B model with a sequence length of 256 and batch size of
256. We benchmark different configurations of Sparse Wide GPT-3 20B with
sparsity $\in$ \{50\%, 75\%, 90\%\} and report seconds/ iteration. 

\paragraph{Benchmarking Analysis} Figure~\ref{fig:sparsity_benchmark} in
Section~\ref{sec:wall_clock} presents the results of benchmarking inference and
training of Sparse-IFT Sparse Wide family. In both setups, we measure the
relative increase in latency or training speed for Sparse-IFT variants against
the dense model. Note that configurations of Sparse-IFT at different values of
sparsity do not incur a significant change in the FLOPs compared to the dense
model. On ideal hardware, FLOPs should translate directly to wall clock time,
and hence, the inference latency or training time for all configurations of
Sparse-IFT should be the same as that of the dense model (dotted black line).
Conversely, when hardware does not support unstructured sparsity, the latency or
training time of Sparse-IFT variants increases with sparsity (blue line). 

The results in Figure~\ref{fig:sparsity_benchmark} of
Section~\ref{sec:wall_clock} show that up 75\%, there is minimal computational
overhead compared to training the original dense baseline model. At 90\%
sparsity, our results lie between these two spectrums (green line). Using Neural
Magic's sparse inference runtime, we observe a significant reduction in
inference latency, bringing down the relative increase in latency from 19.5x to
3.5x. Similiarly, in the case of training on the Cerebras CS-2, we observe a
significant reduction in training-time, bringing down the relative increase from
10.6x to 2.8x. The dense GPT-3 model achieved a throughput of 828.71 iterations
per second on the CS-2, while the Sparse Wide IFT variants recorded throughputs
of 637.5, 595.3, and 294.3 at respective sparsity levels, resulting in overheads
of 1.30x, 1.39x, and 2.82x, respectively.

In Figure~\ref{fig:matmulspeedup}, we illustrate the achievable benefits of
unstructured weight sparsity when utilizing specialized hardware designed for
deep learning, such as the Cerebras CS-2. This figure was regenerated based on
the plot in~\citep{lie_2021}.

\section{Author Contributions} \label{app:authorcontrib}

We provide a summary of each author’s contributions:

\begin{itemize} \item Vithursan Thangarasa was an integral part of the project
    by participating in discussions with Shreyas Saxena and contributing to the
    method. He also implemented all Sparse-IFT transformations in PyTorch,
    proposed using non-linearity in Sparse-IFT, analyzed DST methods via
    spectral analysis, conducted experiments for the entire study on CIFAR-100
    and its ablations, obtained initial results on ImageNet, extended Sparse-IFT
    to efficient architectures (e.g., BotNet, MobileViT), and contributed to
    writing several sections of the manuscript.

    \item Shreyas Saxena conceived the key idea of matching the FLOPs of Sparse
    Wide transformation to a compact dense model, extended the idea to other
    members of the Sparse-IFT family, helped with the implementation,
    established cardinality of Sparse-IFT members to explain the results,
    benchmarked Sparse-IFT for inference, and wrote several sections of the
    manuscript.

    \item Abhay Gupta validated sparse optimizers in PyTorch, conducted
    experiments with Sparse-IFT ResNet variants on ImageNet, helped with
    pre-training of Sparse-IFT variants of GPT on Cerebras CS-2, conducted all
    experiments of Sparse-IFT on downstream CV tasks, and contributed to writing
    parts of the manuscript.
    
    \item Sean Lie helped with the bring-up of sparsity support on Cerebras CS-2
    which was crucial for benchmarking and training Sparse-IFT variants of GPT
    models, and provided feedback to improve the structuring and presentation of
    the manuscript.
    
\end{itemize}


\end{document}